\documentclass[preprint,12pt, authoryear]{elsarticle}

\usepackage[a4paper, left=2.5cm, right=2.5cm, top=2.5cm, bottom=2.5cm]{geometry}

\usepackage{amsmath, amssymb, amsfonts, amsthm}

\usepackage{booktabs}
\usepackage{array}
\usepackage{arydshln}
\usepackage{siunitx}
\usepackage{multirow}

\usepackage{graphicx}
\usepackage{subcaption}
\usepackage[inkscapeformat=pdf]{svg}

\usepackage{xcolor}
\usepackage{todonotes}

\usepackage[T1]{fontenc}
\usepackage{lmodern}

\usepackage{rotating}
\usepackage{float}
\usepackage{multicol}
\usepackage{xspace}
\usepackage{enumitem} % in the preamble

\usepackage[ruled,vlined, linesnumbered]{algorithm2e}
\definecolor{softgreen}{RGB}{180,255,180}
\usepackage[normalem]{ulem} % 

\journal{Computational Statistics \& Data Analysis}

\begin{document}
\synctex=1
\begin{frontmatter}

\title{A new classification method based on Minimum Spanning Trees}

\author[1,2]{Julio González-Díaz}
\author[1,2]{Beatriz Pateiro-López}
\author[1,2]{Iria Rodríguez-Acevedo}

\affiliation[1]{organization={Department of Statistics, Mathematical Analysis and Optimization and MODESTYA Research Group}, 
            addressline={University of Santiago de Compostela}, 
            city={Santiago de Compostela},
            postcode={15782}, 
            state={Galicia},
            country={Spain}}

\affiliation[2]{organization={CITMAga (Galician Center for Mathematical Research and Technology)}, 
            addressline={University of Santiago de Compostela}, 
            city={Santiago de Compostela}, 
            postcode={15782}, 
            state={Galicia}, 
            country={Spain}}
%% Abstract
\begin{abstract}
Minimum Spanning Trees have been used in unsupervised learning, particularly in clustering tasks, due to their ability to recognize clusters by removing edges that are considered inconsistent in defining those clusters. This paper aims to study the use of Minimum Spanning Trees in supervised learning. Specifically, we propose a classification algorithm based on Minimum Spanning Trees. To improve its performance, we introduce a robust version of the method that is also computationally more efficient. We evaluate the effectiveness of our proposed method through an extensive simulation study. We also apply the proposed methodology to a real-world case study involving aircraft trajectories.
\end{abstract}



%% Keywords

\end{frontmatter}



\section{Introduction}

A Minimum Spanning Tree (MST) is a subgraph of a connected, weighted, undirected graph that connects all vertices without forming any cycles, while minimizing the total edge weight. The Minimum Spanning Tree Problem (MSTP) consists of finding such a tree for a given graph. Its importance and popularity arise from its numerous applications and its central role in combinatorial optimization; see \citet{gra85} for a historical overview of the problem. An important advantage of MSTs is their straightforward construction, made possible by efficient greedy algorithms such as those developed by \citet{kruskal} and \citet{prim}. These algorithms make locally optimal choices at each stage and result in a globally optimal solution. The availability of such efficient methods makes the MSTP especially practical for large-scale graphs, which largely explains its broad applicability.  

MSTs are used in many practical settings where network structures need to be designed or analyzed, including areas such as computer and communication systems, transportation networks, etc. MSTP algorithms are also used inside larger optimization tasks, including the traveling salesman problem, multiterminal flow problem, or the matching problem. Because of these and other applications, the MSTP has been the subject of extensive theoretical and algorithmic research. A comprehensive survey can be found in \citet{bazlamaccci}.

In addition to their many practical applications, MSTs have also been extensively studied from a probabilistic perspective. A large body of work focuses on MSTs on random vertex sets in Euclidean space, investigating the stochastic behavior of functionals such as the sum of the power-weighted edge lengths or the number of vertices of a given degree. Main results include laws of large numbers, central limit theorems, and convergence rates; see the book by \citet{yuk98} and the subsequent works by \citet{lee00}, \citet{yuk00}, and \citet{pen03}. More recently, \citet{gan21} studied the asymptotic behavior of MST on a random geometric graph, in which the edges connect points within a given distance threshold.

In statistical learning, MSTs have been primarily used in unsupervised methods, particularly for clustering. Following the classical approach of \citet{zah71}, standard MST-based clustering methods represent the dataset as a graph, where each data point is a node and edges correspond to pairwise distances. An MST is first constructed over the entire dataset, and then the inconsistent edges (for example, those with unusually large weights) are repeatedly removed to form connected components until a terminating criterion is met, with the resulting connected components taken as the clusters. MST-based clustering has been extensively applied in various domains, such as biological data analysis \citep{xu01} and image processing \citep{sa17, li20}, among others. We refer to \citet{gag25} for a review on several MST-based clustering methods and a comparative study of their performance in partitional clustering.

By contrast, the application of MSTs to supervised learning has been far less explored. Supervised learning aims to predict the class or value of new, unseen instances using a model trained on labeled data. In classification, for example, the goal is to assign each input to one of several predefined classes based on its features. Numerous algorithms have been proposed, including Decision Trees, Support Vector Machines, $k$-Nearest Neighbors, and Neural Networks. Although MSTs have been widely used in unsupervised learning and other applications, they have seldom been employed to define classification rules. Their ability to capture global structural properties of the data suggests that they may provide a useful alternative to traditional classification approaches in certain scenarios, particularly in problems where the data exhibit specific structural patterns, such as network-related data.

Motivated by these considerations, we propose a method that constructs an MST for each class within a given training dataset and assigns a new observation to the class whose MST is least affected by its inclusion. The effect of this inclusion is quantified using a metric called the conformity of each class, which measures the change in the average cost per vertex of the MST after adding the new observation. A higher conformity indicates a stronger affinity with the class, thus guiding the classification decision.  In order to assess its performance, we demonstrate the potential of the proposed approach through detailed numerical experiments and a real-world case study.

Coinciding with our work, recent independent studies by  \citet{mst1} and \citet{mst2}  have also explored MST-based classifiers. While this highlights an increasing interest in the topic, our framework differs in several key aspects. Rather than minimizing absolute network cost changes, our method evaluates a normalized ratio of MST variation, providing more balanced comparisons across uneven class sizes. In addition, we incorporate mechanisms to handle mislabeled data and improve computational efficiency, which are core components of our method. Furthermore, to complement previous findings, our research offers an in-depth computational analysis, focusing on the underlying network patterns. We aim to provide a more thoroughly validated framework that addresses previously unexplored aspects of MST-based classification.

The paper is organized as follows. In Section~\ref{Section:MST}, we introduce the proposed MST-based classifier. Section~\ref{Section:MST-RClass} presents a robust extension that improves performance in the presence of mislabeled data and increases computational efficiency. In Section~\ref{Section:numerical-study}, we describe the computational setup and provide an extensive simulation study, as well as an application to real-world data. Finally, Section~\ref{Section:conclusions} presents some concluding remarks and outlines the combinatorial challenges of establishing theoretical guarantees for the proposed methods.

\section{MST classifier}\label{Section:MST}

We introduce here a classification method based on MSTs, which we will denote as \mbox{MST-Class}.  For simplicity, we first formalize the method in a binary classification problem with input vectors \(X=(x_1,\dots,x_d)\in\mathbb{R}^d\) and labels \(Y\in\mathcal{Y}=\{0,1\}\).  
We observe a training sample of $n$ independent and identically distributed (i.i.d.) pairs $(X_1,Y_1),\dots,(X_n,Y_n)$, with the same distribution as $(X,Y)$. Let $n_0$ and $n_1$ denote the number of observations such that $Y_i = 0$ and $Y_i = 1$, respectively, for $i \in \{1, \dots, n\}$.  

\begin{figure}[h!]
    	\centering
      \includegraphics[width=0.45\linewidth]{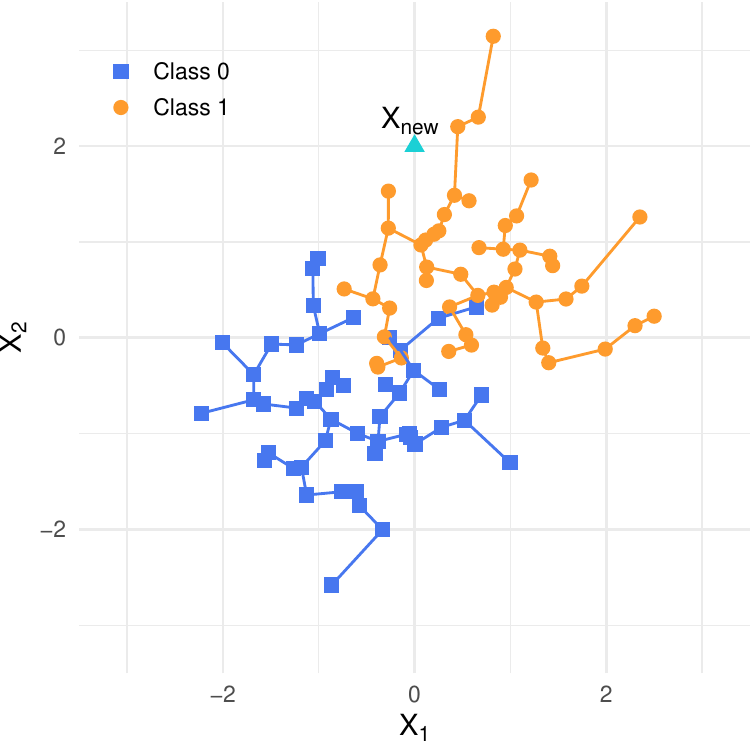}
      \includegraphics[width=0.45\linewidth]{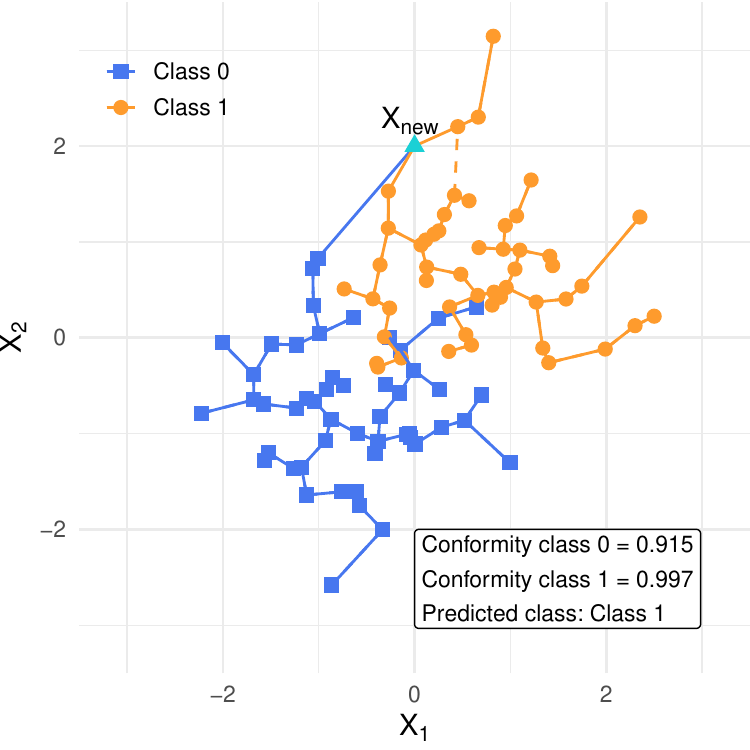}
        \caption{Illustration of Algorithm ~\ref{alg_MST} (MST-Class). Left: MSTs of each class computed from the training data, together with the new observation $X_{\rm new}$. Right: MSTs after adding $X_{\rm new}$ to each class (edges that change are shown as dashed lines). Since $\mathrm{Conf}_1 > \mathrm{Conf}_0$, $X_{\rm new}$ is assigned to class ~1.}
        \label{fig:MST_Alg1}
\end{figure}

The main idea is to represent the observations in each class as a weighted graph that captures the internal similarity among its observations.  Throughout the exposition we use the Euclidean distance as the edge weight, although other distances or dissimilarity measures could also be employed. Recall that, given a finite set of points $V=\{v_1,\dots,v_m\}\subset\mathbb{R}^d$, its (Euclidean) MST is defined as the spanning tree $T(V)$ on the complete graph with vertex set \(V\) that minimizes the sum of its edge lengths, that is,
\[
\sum_{e\in T(V)} \|e\|  
\;=\; 
\min_{T'} \left\{ \sum_{e\in T'} \|e\| : T' \text{ is a spanning tree on } V \right\},
\]
where an edge \(e=(v_i,v_j)\) has weight the Euclidean distance between its endpoints {\mbox{\(\|v_i - v_j\|\)}}.  Thus, we compute the MST for each class. We then define the \emph{separation} of a class as the average edge length of its MST, that is, the total length (cost) divided by the number of observations in the class. This provides a measure of how compact or tightly grouped the observations are within the class. Once the MSTs of the two classes have been computed, we consider a new observation. To evaluate how well the new observation fits each class, we compute again the MST for each class  including the new point and calculate the corresponding separation. Comparing the separation before and after adding the point gives the so-called \emph{conformity} for each class. Formally, let $\mathrm{Sep}_j$ denote the separation of class $j$ before adding the new observation, and let $\mathrm{Sep}_j^{\rm new}$ be the separation after including the observation. 
The conformity of class $j$ is then defined as
\[
\mathrm{Conf}_j = \frac{\mathrm{Sep}_j}{\mathrm{Sep}_j^{\rm new}}.
\]
A higher conformity indicates that the new observation fits well with the existing points in the class, as it reduces or minimally increases the average MST length. Finally, the classification rule assigns the new observation to the class for which the conformity is highest.  The full procedure is detailed in Algorithm~\ref{alg_MST}. Figure~\ref{fig:MST_Alg1} illustrates the method visually.

\begin{algorithm}[!ht]
\caption{MST-Class: Minimum Spanning Tree Classifier}
\label{alg_MST}
\KwIn{Training data $\mathcal{D}_n=\{(X_i,Y_i)\}_{i=1}^n\subset\mathbb{R}^d\times\{0,1\}$ and new observation $X_{\text{new}}$.}
\KwOut{Predicted class $\hat{Y}_{\rm new} = g_n(X_{\rm new})$.}
\BlankLine
\BlankLine
Standardize the training data and apply the same transformation to $X_{\rm new}$.

    \ForEach{class $j \in \{0,1\}$}{
        Build the complete graph on its points using Euclidean distances as edge weights.
        
        Compute the MST.

        Let $n_j$ be the class size and $C_j$ be the total MST cost. Compute the separation $\mathrm{Sep}_j = C_j / n_j.$
        
    Insert the standardized observation $X_{\rm new}$ into the class graph and recompute the MST.
    
    Let $C_j^{\rm new}$ be the updated MST cost.    Compute $\mathrm{Sep}_j^{\rm new} = C_j^{\rm new} / (n_j + 1)$.
    
    Compute the conformity
    \[
\mathrm{Conf}_j = \frac{\mathrm{Sep}_j}{\mathrm{Sep}_j^{\rm new}}.
\]
}

 \KwRet{predicted class: $\displaystyle
g_n(X_{\rm new}) =
\begin{cases}
1 & \text{if } \mathrm{Conf}_1 > \mathrm{Conf}_0,\\[4pt]
0 & \text{if } \mathrm{Conf}_1 < \mathrm{Conf}_0,\\[4pt]
\text{Randomize} & \text{otherwise}.
\end{cases}$}

\end{algorithm}

The MST-based classification method described in Algorithm~\ref{alg_MST} for the binary setting can be naturally extended to multi-class problems. Suppose that there are $K$ classes. For each class, we can consider the subset of training observations belonging to that class and construct a complete graph in which nodes represent the observations and edges are weighted according to the distances between them. We then compute the MST for each class graph and use its total cost, normalized by the number of observations in the class, as a measure of within-class separation. To classify a new observation, we add it to each class graph and recompute the MSTs to measure its impact on the corresponding separation. The conformity of the observation with each class is quantified as the relative change in separation.  Finally, the observation is assigned to the class with the highest conformity value. As in the binary case, ties are resolved randomly. This multi-class extension follows the same principles as the original binary classification method, applying them simultaneously across all classes.

\section{MST robust classifier}\label{Section:MST-RClass}

\begin{figure}[h!]
    \centering
    \includegraphics[width=0.45\linewidth]{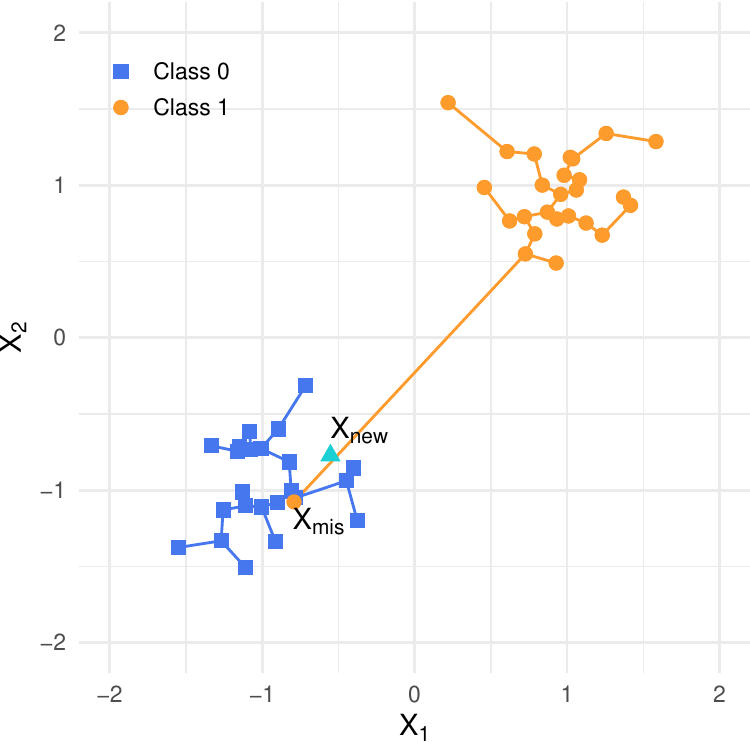}
    \includegraphics[width=0.45\linewidth]{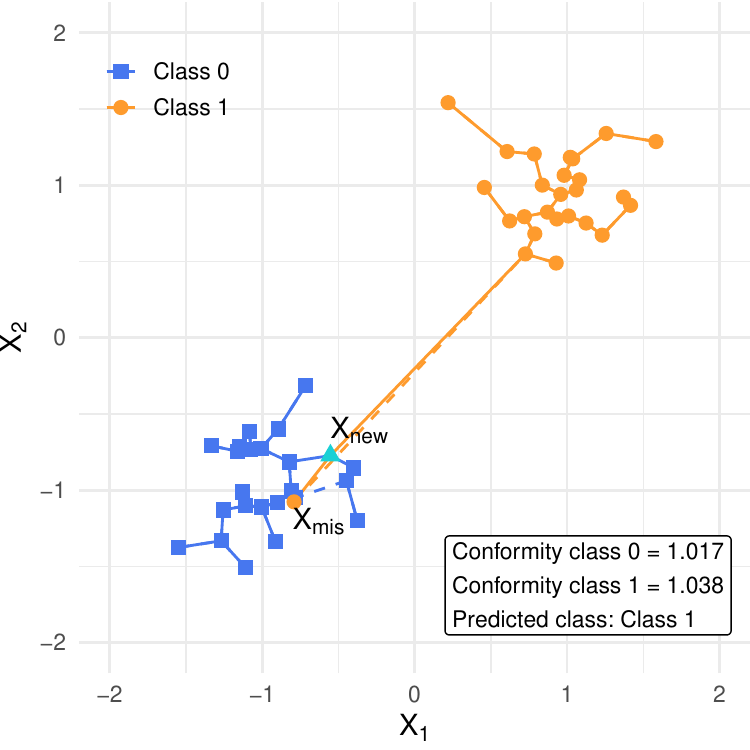}
    \caption{Illustration of Algorithm~\ref{alg_MST} (MST-Class) in the presence of label noise. {Left:} MSTs of each class computed from the training data, including a mislabeled observation $X_{\rm mis}$. {Right:} MSTs after adding $X_{\rm new}$ to each class (edges that change are shown as dashed lines). Despite $X_{\rm new}$ being closer to class 0 in feature space, it is assigned to class 1 due to the influence of the mislabeled point on the separation of the MSTs.}
    \label{fig:example}
\end{figure}

It is well known that the performance of any classification method depends on the quality of the data used for training. Among the factors that can affect data quality, one particularly important aspect is label noise: errors in the class assignments of some training observations. Such mislabeling can significantly disrupt the learning process, causing classifiers to make systematic mistakes. The MST-Class method proposed in Section~\ref{Section:MST} is especially sensitive to this type of noise. To illustrate this, Figure~\ref{fig:example} shows a small toy example where the training set contains a mislabeled point $X_{\rm mis}$, and the MST-Class rule is used to classify a new observation $X_{\rm new}$. Despite $X_{\rm new}$ being closer to class 0, the mislabeled observation $X_{\rm mis}$ leads MST-Class to assign it to class 1, due to its effect on the separation of the corresponding MSTs.  This scenario motivates the development of a robust variant of MST-Class, which we refer to as MST-RClass (MST Robust Classifier). The full procedure is detailed in Algorithm~\ref{alg_MST_R}.

\begin{algorithm}[!ht]
\caption{MST-RClass: Minimum Spanning Tree Robust Classifier}
\label{alg_MST_R}

\KwIn{Training data $\mathcal{D}_n=\{(X_i,Y_i)\}_{i=1}^n\subset\mathbb{R}^d\times\{0,1\}$, new observation $X_{\rm new}$, number of subsamples $s$, and size of each subsample $m$.}
\KwOut{Predicted class $\hat{Y}_{\rm new} = g_n^{\rm R}(X_{\rm new})$.}

\BlankLine
\BlankLine
Standardize the training data and apply the same transformation to $X_{\rm new}$.

\For{$l = 1$ \KwTo $s$}{

    Randomly select $m$ points from the standardized training data.
    
    \ForEach{class $j \in \{0,1\}$}{
        Build the complete graph on its points using Euclidean distances as edge weights.
        
        Compute the MST.
        
        Let $m_j$ be the class size and $C_j$ be the total MST cost. Compute the separation $\mathrm{Sep}_j = C_j / m_j.$
        
    Insert $X_{\rm new}$ into the class graph and recompute the MST.
    
    Let $C_j^{\rm new}$ be the updated MST cost.    Compute $\mathrm{Sep}_j^{\rm new} = C_j^{\rm new} / (m_j + 1)$.
    
    Compute the conformity
    \[
\mathrm{Conf}^{(l)}_j = \frac{\mathrm{Sep}_j}{\mathrm{Sep}_j^{\rm new}}.
\]
    }
    }

\ForEach{class $j \in \{0,1\}$}{  
Compute the average conformity:
\[
  \overline{\mathrm{Conf}}_j = \frac{1}{s}\sum_{l=1}^s
  \mathrm{Conf}_j^{(l)}.
\]
}

 \KwRet{predicted class: $\displaystyle
g_n^{\rm R}(X_{\rm new}) =
\begin{cases}
1 & \text{if } \overline{\mathrm{Conf}}_1 > \overline{\mathrm{Conf}}_0,\\[4pt]
0 & \text{if } \overline{\mathrm{Conf}}_1 < \overline{\mathrm{Conf}}_0,\\[4pt]
\text{Randomize} & \text{otherwise}.
\end{cases}$
}
\end{algorithm}

MST-RClass classifies a new observation by combining multiple subsampled versions of the MST classifier. In each of the $s$ repetitions, a subsample of the training set is drawn, class-specific MSTs are built, and the structural change caused by inserting the new point is measured, yielding a conformity value for each class. The method then averages these conformity scores across all subsamples and assigns the observation to the class with the highest mean conformity. This strategy mitigates the influence of outliers or mislabeled points, providing a more robust alternative to the baseline MST classifier. Returning to the toy example in Figure \ref{fig:example}, many subsamples may exclude the observation $X_{\rm mis}$, resulting in higher conformity of $X_{\rm new}$ in class $0$ than in class $1$. As a consequence, by averaging the conformity values returned by all the subsamples, we would certainly end up doing the classification correctly. 

It is relevant to mention that the choice of size for the subsamples is particularly important because of the order of complexity of the algorithms for the MSTP. The computational complexity of Prim's  or Kruskal's algorithms \citep{prim,kruskal} depends on certain implementation details, which are beyond the scope of this discussion. However, it is never worse than $O(n^2)$, where $n$ is the number of vertices of the graph. This is the value we use as a reference. Thus, in a balanced case in which the classes contain a similar number of observations, calculating the MSTs required for using MST-Class has a complexity of $O(n^2)$. In contrast, if Algorithm~\ref{alg_MST_R} uses subsamples of size $m$, for example $m=\sqrt{n}$, the computational complexity for constructing the corresponding MSTs drops to $O(m^2)=O(n)$. Thus, the computation time changes from quadratic in the number of observations to linear in the number of observations (multiplied by $s$, typically much smaller than $n$ in large datasets). This computational improvement is particularly relevant in big data contexts, where transitioning from quadratic to linear complexity has a significant impact. In the subsequent computational study, we examine in greater detail the impact of this procedure on selected particular cases, thereby providing a clearer view of its benefits, not only in terms of improved classification accuracy but also in reduced computational time.

\section{Numerical study}\label{Section:numerical-study}

The computational study is structured in two main parts, as detailed in Subsections~\ref{Section:simulated-data} and~\ref{Section:trajectories}. In the first part (Subsection~\ref{Section:simulated-data}), we conduct a comprehensive comparison between MST-Class (Algorithm  \ref{alg_MST}) and MST-RClass (Algorithm  \ref{alg_MST_R}), evaluating their performance both against each other and against several widely used classification techniques on simulated datasets. In the second part (Subsection~\ref{Section:trajectories}), we present a case study based on real aircraft trajectory data, where the underlying distribution of the data is unknown and potentially more challenging.

Before presenting the results, we describe in Subsection~\ref{sec:env} the computational setup employed in this study.

\subsection{Testing environment}\label{sec:env}
All computations were performed on the Finisterrae III supercomputer at the Galicia Supercomputing Centre (CESGA). Execution time and memory allocation per node were adapted to the size of each dataset, with smaller datasets requiring fewer resources. All classification methods were implemented in R, see \citet{R}. Random seeds were set to ensure reproducibility, and reported execution times correspond to wall-clock (elapsed) time. MSTs were calculated using the \textit{igraph} package, employing Prim's algorithm via its \textit{mst} function, see \citet{igraph}. We compare the proposed MST-based classifiers with standard alternatives, including $k$-Nearest Neighbors (KNN), Linear Discriminant Analysis (LDA), and Quadratic Discriminant Analysis (QDA), the latter two being included because they are theoretically optimal for normally distributed data. All three classifiers are implemented using the \textit{caret} package, see \citet{caret}.

\subsection{Experiments on simulated data}\label{Section:simulated-data}

We conduct a comprehensive analysis on simulated datasets, which allows us to systematically explore a wide range of configurations, dimensions, and sample sizes. We examine the effect of factors such as class separation and label noise, providing a detailed evaluation of each method under controlled conditions.

Two main types of scenarios are considered. The first consists of multivariate normal data, which follow a standard statistical structure and do not provide any particular advantage to MST-based classifiers. The second involves datasets with a strong underlying geometric organization, where MST-based classifiers may be more competitive than standard methods. 

\subsubsection{Normal data}\label{Section:normal-data}

We begin by considering a binary classification problem, where each class is generated from a predefined multivariate normal distribution. This scenario provides a standard statistical setting, allowing us to evaluate the performance of the classifiers under classical distributional assumptions.

\begin{figure}[!htbp]
	\centering
	\includegraphics[scale=0.8]{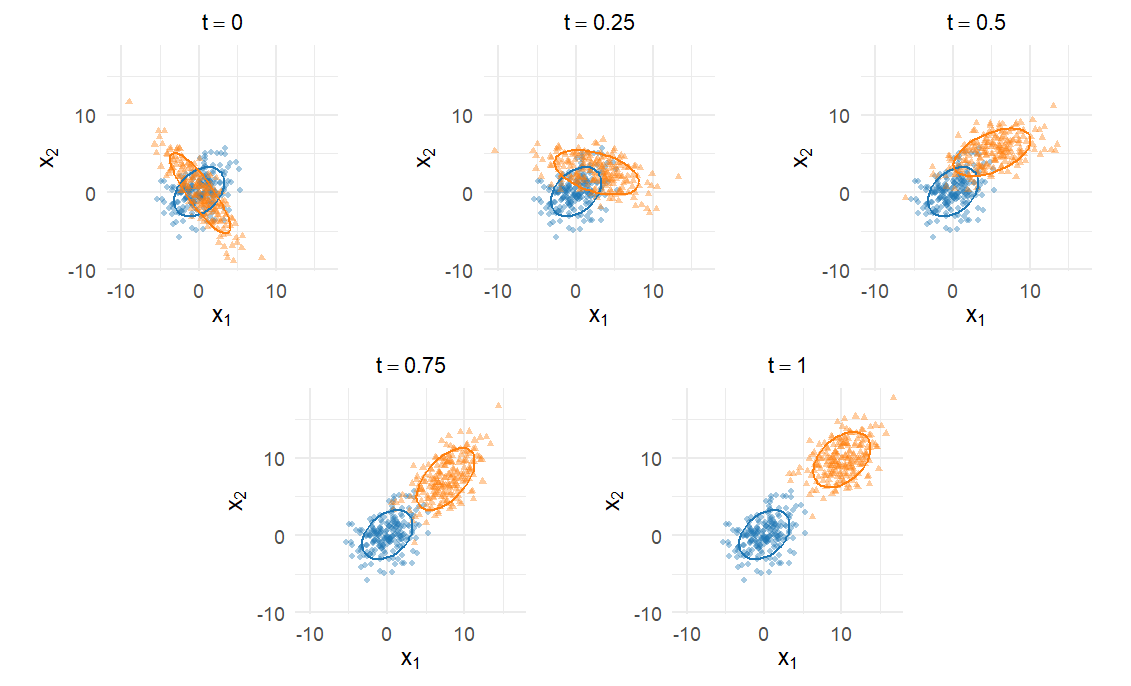}
	\caption{Bivariate normal datasets for two classes in $\mathbb{R}^2$, shown at selected stages of the second class evolution ($t\in\{0, 0.25, 0.5, 0.75, 1\}$). The first class remains fixed, while the second class gradually changes its mean and covariance.}
	\label{fig:norm_dim2_without}
\end{figure}

We first focus on the bidimensional case $d=2$, which additionally allows us to visualize the resulting class configurations. The first class is kept fixed in both mean and covariance. The second class evolves along a smooth trajectory: its mean changes linearly while its covariance gradually interpolates from an initial configuration toward that of the first class, as a function of $t \in [0,1]$. Detailed specifications of the procedure used to generate the datasets, including the specific parameter values, are provided in the Supplementary Material. To illustrate the effects of this dynamic evolution, we consider five data configurations corresponding to $t\in\{0, 0.25, 0.5, 0.75, 1\}$. A representation of these configurations can be found in Figure \ref{fig:norm_dim2_without}, where the first class is represented in circles and the second one in triangles. The parameter $t$ thus serves as a clear indicator of classification difficulty: low $t$ corresponds to challenging scenarios, while high $t$ corresponds to easily separable classes.

For each configuration and each class sample size $n_0 = n_1 \in\{ 250, 500, 1000\}$, we generate 100 datasets. Within each dataset, observations are randomly split into training and test sets, with 70\% used for training and 30\% for testing. The reported results correspond to the average classification accuracy on the test sets across all 100 datasets, with the standard deviation indicated in parentheses. For the KNN classifier, the tuning parameter $k$ is chosen to maximize accuracy via repeated 10-fold cross-validation. The results are summarized in  Table \ref{tab:two-norm-all}. We observe that the classification accuracy increases consistently with $t$, reflecting that as the second class gradually shifts its mean along the diagonal direction and becomes more aligned in covariance with the first class, the task becomes easier.   While  MST-Class does not achieve the highest accuracy, it maintains a stable performance across configurations, comparable in many cases to that of KNN. These results also corroborate theoretical expectations:  QDA outperforms  LDA when covariances differ, whereas  LDA achieves optimal performance when covariances are equal.

\begin{table}[!htbp]
\centering
\begin{tabular}{c l c c c c}
\toprule
Class size & $t$ & MST-Class & KNN & LDA & QDA \\
\midrule
$n_0=n_1=250$ & $0$ & 0.688 (0.039) & 0.736 (0.039) & 0.492 (0.058) & 0.754 (0.036) \\
               & $0.25$ & 0.719 (0.038) & 0.801 (0.031) & 0.789 (0.032) & 0.808 (0.032) \\
               & $0.5$ & 0.804 (0.040) & 0.891 (0.027) & 0.899 (0.027) & 0.899 (0.027) \\
               & $0.75$ & 0.938 (0.031) & 0.967 (0.019) & 0.969 (0.015) & 0.969 (0.016) \\
               & $1$ & 0.994 (0.007) & 0.995 (0.006) & 0.996 (0.005) & 0.996 (0.005) \\
\midrule
$n_0=n_1=500$ & $0$ & 0.673 (0.034) & 0.743 (0.028) & 0.499 (0.053) & 0.758 (0.027) \\
               & $0.25$ & 0.707 (0.030) & 0.805 (0.023) & 0.794 (0.025) & 0.816 (0.021) \\
               & $0.5$ & 0.799 (0.029) & 0.893 (0.018) & 0.896 (0.017) & 0.897 (0.017) \\
               & $0.75$ & 0.932 (0.020) & 0.966 (0.011) & 0.967 (0.011) & 0.968 (0.010) \\
               & $1$ & 0.993 (0.007) & 0.996 (0.004) & 0.997 (0.003) & 0.996 (0.004) \\
\midrule
$n_0=n_1=1000$ & $0$ & 0.668 (0.021) & 0.744 (0.019) & 0.496 (0.032) & 0.754 (0.017) \\
                & $0.25$ & 0.698 (0.021) & 0.810 (0.014) & 0.794 (0.014) & 0.815 (0.014) \\
                & $0.5$ & 0.788 (0.022) & 0.895 (0.014) & 0.897 (0.014) & 0.898 (0.014) \\
                & $0.75$ & 0.923 (0.021) & 0.968 (0.008) & 0.968 (0.008) & 0.968 (0.008) \\
                & $1$ & 0.992 (0.005) & 0.996 (0.003) & 0.996 (0.003) & 0.996 (0.003) \\
\bottomrule
\end{tabular}
\caption{Classification accuracy (mean and standard deviation) for MST-Class, KNN, LDA, and QDA across two-dimensional normal data configurations. Results are averaged over 100 datasets for each combination of configuration $t$ and class sample size.}
\label{tab:two-norm-all}
\end{table}

\subsubsection*{Mislabeled data}
For the same scenarios analyzed previously, we compare MST-Class with MST-RClass. Specifically, for each of the five configurations and each class sample size, 2.5\% of the labels were randomly flipped. This setup allows us to evaluate how each method handles label noise and to determine whether MST-RClass, which is specifically designed to be more robust, maintains higher accuracy than the standard MST-Class under these conditions.

As described in Algorithm \ref{alg_MST_R}, MST-RClass depends on two main parameters: the number of subsamples, $s$, and the size of each subsample, $m$. In our experiments, we have fixed $s=100$, which we consider sufficiently large to provide a representative view of the dataset. Also, although Algorithm~\ref{alg_MST_R} is presented in terms of subsamples of total size $m$, in practice we construct each subsample in a class-balanced way. That is, instead of drawing $m$ observations from $\mathcal{D}_n=\{(X_i,Y_i)\}_{i=1}^n$ directly, we sample  $m_0$ observations from class 0 and $m_1$ observations from class 1, typically taking $m_0=m_1$, so that each subsample contains the same number of points from each class.

In our experiments, we investigate several subsample-size configurations by drawing class-balanced subsamples with $m_0=m_1=\sqrt{n_{\rm min}}$ (MST-RClass1), $m_0=m_1=2\sqrt{n_{\rm min}}$ (MST-RClass2), $m_0=m_1=5\sqrt{n_{\rm min}}$ (MST-RClass5), and $m_0=m_1=10\sqrt{n_{\rm min}}$ (MST-RClass10), where $n_{\rm min}=\min(n_0,n_1)$.  This setup allows us to study the effect of subsample size on both classification accuracy and computational efficiency.

\begin{figure}[!htbp]
	\centering
	\includegraphics[width=0.75\textwidth]{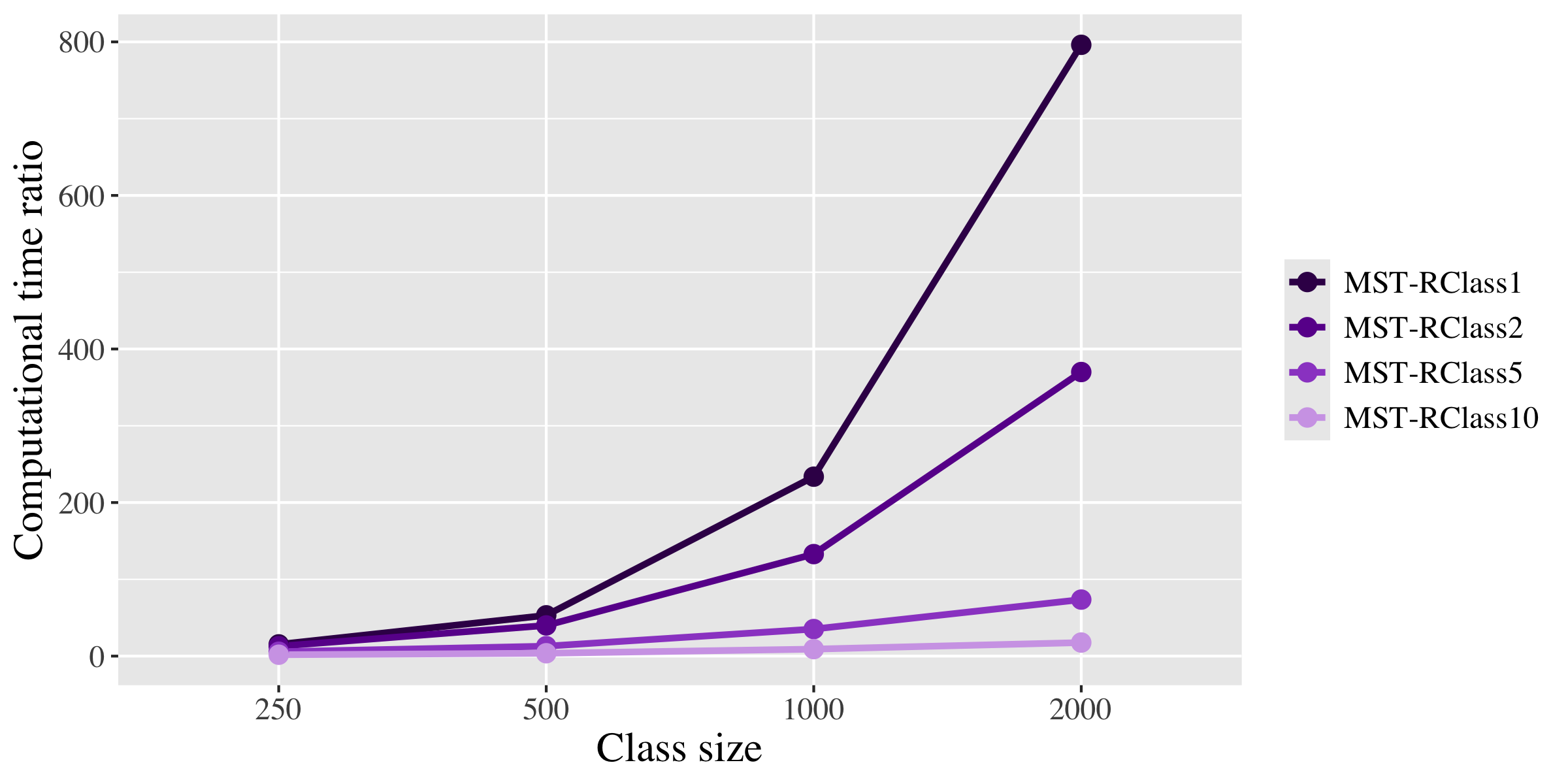}
	\caption{Average computational time ratio $\mathrm{Time(MST\text{-}Class)}/\mathrm{Time(MST\text{-}RClass,\ }s=1)$ across five datasets corresponding to different values of $t$, for class sample sizes $n_0=n_1\in\{250,500,1000,2000\}$ and different MST-RClass configurations.}
	\label{fig:ratios}
\end{figure}

Before discussing classification accuracy in detail, we first compare MST-Class and MST-RClass in terms of computational time. For this experiment, we set $s=1$, which implies that the computational times reported for MST-RClass should be multiplied by a factor of 100 to obtain the corresponding times when $s=100$. Figure~\ref{fig:ratios} shows the average ratio
\[
\frac{\mathrm{Time(MST\text{-}Class)}}{\mathrm{Time(MST\text{-}RClass,\ }s=1)},
\]
computed over five datasets (one for each value of $t$), across four class sample sizes $n_0=n_1 \in \{250,500,1000,2000\}$ and for each MST-RClass configuration. It can be readily seen that MST-RClass configurations scale much better than MST-Class. In particular they show that, for sample sizes 1000 and 2000, MST-RClass1 and MST-RClass2 would already be less demanding computationally than MST-Class even when $s=100$. These scaling results are in agreement with the discussion in Section~\ref{Section:MST-RClass}.

%Since the actual computational time of MST-RClass when $s=100$ is obtained by multiplying the reported times by 100, MST-RClass would outperform MST-Class in absolute computational time when the displayed ratio exceeds 100. This occurs for MST-RClass1 and MST-RClass2 for class sizes of 1000 observations and above.

%The time ratio consistently increases as the sample size increases for all subsample sizes. This confirms that, despite its higher initial cost, MST-RClass scales more efficiently with the number of observations, in agreement with the discussion in Section~\ref{Section:MST-RClass}.

Table~\ref{tab:mstclass_vs_mstrclass_combined_all} reports the accuracy results for $n_0=n_1\in\{250, 500,1000\}$, respectively. We observe that MST-RClass, regardless of the original number of observations per class and the selected subsample size, consistently outperforms  MST-Class across all configurations. This results in significantly improved classification performance, particularly when the separation between classes is larger (higher values of $t$). In these scenarios, mislabeled observations are more easily distinguished, whereas for lower values of $t$, when the classes are more overlapping, there is a higher risk of misclassification. For instance, when $n_0=n_1=500$, for $t = 1$, the accuracy increases from 0.749 with  MST-Class to 0.970 when using  {MST-RClass1}.

\begin{table}[!htbp]
	\centering
	\resizebox{\textwidth}{!}{%
	\begin{tabular}{c l c c c c c}
		\toprule
		Class size & $t$ & MST-Class & MST-RClass1 & MST-RClass2 & MST-RClass5 & MST-RClass10 \\
		\midrule
		$n_0=n_1=250$ & $0$ & 0.656 (0.043) & 0.737 (0.038) & 0.732 (0.038) & 0.716 (0.041) & 0.684 (0.038) \\
		& $0.25$ & 0.667 (0.042) & 0.791 (0.034) & 0.788 (0.034) & 0.762 (0.035) & 0.710 (0.036) \\
		& $0.5$ & 0.714 (0.053) & 0.876 (0.028) & 0.871 (0.029) & 0.850 (0.030) & 0.788 (0.042) \\
		& $0.75$ & 0.763 (0.059) & 0.940 (0.019) & 0.938 (0.020) & 0.929 (0.021) & 0.865 (0.037) \\
		& $1$ & 0.771 (0.062) & 0.966 (0.014) & 0.966 (0.014) & 0.960 (0.013) & 0.877 (0.037) \\
		\midrule
		$n_0=n_1=500$ & $0$ & 0.639 (0.038) & 0.740 (0.027) & 0.738 (0.027) & 0.719 (0.032) & 0.694 (0.031) \\
		& $0.25$ & 0.659 (0.030) & 0.796 (0.022) & 0.789 (0.023) & 0.768 (0.024) & 0.734 (0.026) \\
		& $0.5$ & 0.705 (0.036) & 0.872 (0.018) & 0.870 (0.019) & 0.860 (0.018) & 0.827 (0.024) \\
		& $0.75$ & 0.745 (0.053) & 0.941 (0.014) & 0.941 (0.014) & 0.936 (0.014) & 0.908 (0.021) \\
		& $1$ & 0.749 (0.054) & 0.970 (0.008) & 0.969 (0.008) & 0.962 (0.009) & 0.930 (0.020) \\
		\midrule
		$n_0=n_1=1000$ & $0$ & 0.627 (0.024) & 0.735 (0.017) & 0.730 (0.017) & 0.718 (0.018) & 0.703 (0.019) \\
		& $0.25$ & 0.647 (0.024) & 0.796 (0.015) & 0.792 (0.015) & 0.777 (0.018) & 0.754 (0.020) \\
		& $0.5$ & 0.702 (0.026) & 0.874 (0.015) & 0.871 (0.014) & 0.862 (0.015) & 0.846 (0.016) \\
		& $0.75$ & 0.745 (0.032) & 0.946 (0.009) & 0.945 (0.009) & 0.942 (0.010) & 0.931 (0.011) \\
		& $1$ & 0.763 (0.033) & 0.971 (0.006) & 0.970 (0.006) & 0.965 (0.007) & 0.956 (0.011) \\
		\bottomrule
	\end{tabular}
}
	\caption{Comparison of MST-Class and MST-RClass across configurations of two-dimensional normal data with mislabeled points, for 250, 500, and 1000 observations per class.}
	\label{tab:mstclass_vs_mstrclass_combined_all}
\end{table}

The results also show a clear effect of subsample size on the performance of  MST-RClass. Smaller subsamples consistently achieve higher classification accuracy across all configurations of $t$. This behavior is intuitive: as the subsample size increases, MST-RClass increasingly resembles MST-Class, reducing its robustness to label noise due to the inclusion of mislabeled observations. Nevertheless, in some cases, using a slightly larger subsample than in {MST-RClass1} can be beneficial, as shown in later analyses. Depending on the underlying data structure, constructing the MST with a subsample size that is neither too small nor as large as the full dataset can improve classification performance, effectively balancing robustness to mislabeled observations with the ability to capture the overall structure of the data. This leads us to implement a cross-validation procedure for selecting the subsample size in MST-RClass. The training set is used to choose an appropriate subsample size before classifying the test set, allowing the method to adapt to the characteristics of each dataset and balance accuracy with robustness. This approach keeps the subsample large enough to capture meaningful patterns while small enough to maintain resistance to label noise, resulting in improved performance across diverse scenarios.

The candidate subsample sizes considered in this cross-validation procedure are multiples of $\sqrt{n_{\rm min}}$, specifically $h\sqrt{n_{\rm min}}$ with $h \in \{1, 2, 5, 10\}$. Using this procedure, we re-evaluate the performance of  MST-Class and  MST-RClass with a cross-validated subsample size, comparing them to the other classification methods:  KNN,  LDA, and  QDA. Table~\ref{tab:two-norm-all-mis} presents the results, including two additional columns: the mean and standard deviation of the multiplying factor $h$ that determines the subsample size for MST-RClass, and the mean and standard deviation of the number of neighbors $k$ selected for KNN, both determined via cross-validation.

\begin{table}[!htbp]
	\centering
	\resizebox{\textwidth}{!}{%
		\begin{tabular}{c l c cc cc c c}
			\toprule
			Class size & $t$ & MST-Class & \multicolumn{2}{c}{MST-RClass} & \multicolumn{2}{c}{KNN} & LDA & QDA \\
			\cmidrule(lr){4-5} \cmidrule(lr){6-7}
			&  &  & Accuracy & $h$ & Accuracy & $k$ &  &  \\
			\midrule
			$n_0=n_1=250$ & $0$ & 0.656 (0.043) & 0.730 (0.043) & 2.370 (2.304) & 0.722 (0.040) & 13.720 (5.081) & 0.495 (0.056) & 0.739 (0.039) \\
			& $0.25$ & 0.667 (0.042) & 0.782 (0.039) & 2.170 (1.787) & 0.784 (0.031) & 14.710 (4.468) & 0.772 (0.035) & 0.793 (0.033) \\
			& $0.5$ & 0.714 (0.053) & 0.873 (0.031) & 1.590 (1.045) & 0.871 (0.026) & 14.750 (4.284) & 0.879 (0.027) & 0.879 (0.028) \\
			& $0.75$ & 0.763 (0.059) & 0.939 (0.019) & 1.860 (1.349) & 0.939 (0.020) & 12.870 (5.088) & 0.941 (0.018) & 0.942 (0.018) \\
			& $1$ & 0.771 (0.062) & 0.965 (0.014) & 1.530 (1.210) & 0.966 (0.014) & 14.980 (5.420) & 0.967 (0.013) & 0.967 (0.014) \\
			\midrule
			$n_0=n_1=500$ & $0$ & 0.639 (0.038) & 0.734 (0.030) & 2.140 (1.959) & 0.729 (0.029) & 20.770 (5.876) & 0.493 (0.048) & 0.744 (0.027) \\
			& $0.25$ & 0.659 (0.030) & 0.790 (0.026) & 1.710 (1.465) & 0.788 (0.023) & 20.710 (6.376) & 0.776 (0.024) & 0.798 (0.023) \\
			& $0.5$ & 0.705 (0.036) & 0.870 (0.019) & 1.800 (1.450) & 0.873 (0.018) & 21.050 (6.029) & 0.875 (0.019) & 0.876 (0.019) \\
			& $0.75$ & 0.745 (0.053) & 0.941 (0.013) & 2.100 (1.673) & 0.941 (0.013) & 16.890 (7.047) & 0.943 (0.013) & 0.943 (0.013) \\
			& $1$ & 0.749 (0.054) & 0.969 (0.009) & 1.650 (1.431) & 0.970 (0.008) & 19.350 (8.030) & 0.970 (0.008) & 0.970 (0.008) \\
			\midrule
			$n_0=n_1=1000$ & $0$ & 0.628 (0.024) & 0.732 (0.017) & 1.856 (1.414) & 0.730 (0.018) & 28.670 (9.078) & 0.496 (0.039) & 0.741 (0.017) \\
			& $0.25$ & 0.646 (0.023) & 0.793 (0.016) & 1.571 (0.956) & 0.793 (0.015) & 31.451 (6.729) & 0.777 (0.014) & 0.798 (0.015) \\
			& $0.5$ & 0.702 (0.026) & 0.874 (0.015) & 1.719 (1.312) & 0.875 (0.014) & 30.271 (7.751) & 0.877 (0.015) & 0.878 (0.015) \\
			& $0.75$ & 0.743 (0.033) & 0.945 (0.010) & 1.809 (1.306) & 0.945 (0.009) & 22.819 (10.112) & 0.945 (0.009) & 0.945 (0.010) \\
			& $1$ & 0.762 (0.033) & 0.970 (0.006) & 1.768 (1.207) & 0.971 (0.006) & 20.368 (12.041) & 0.971 (0.005) & 0.971 (0.005) \\
			\bottomrule
		\end{tabular}%
	}
	\caption{Performance metrics for MST-Class, MST-RClass, KNN, LDA, and QDA across configurations of two-dimensional normal data with mislabeled points, for 250, 500, and 1000 observations per class.}
	\label{tab:two-norm-all-mis}
\end{table}

Using this cross-validated version of  MST-RClass, we consistently obtain better results than  MST-Class across all configurations and class sizes. Furthermore, the subsample size selected via cross-validation is typically small, suggesting that, for these datasets, smaller subsamples are preferable. This observation aligns with the conclusions from the previous computational experiment, confirming the effectiveness of the cross-validation procedure. Overall,  MST-RClass tends to achieve slightly higher accuracies than  KNN across most configurations and dataset sizes, although the differences are generally minor. Also,  the selected parameter $k$ in KNN is more sensitive to sample size than $h$ in MST-RClass. 

To conclude the section on normally distributed data, we also explore the behavior of the classifiers in settings with more than two dimensions. In particular, we consider the cases $d=3$ and $d=10$, with no mislabeled data introduced. In both scenarios, the data are generated following procedures analogous to those used in the two-dimensional setting: one class remains fixed while the other varies with~$t$. We focus on the two extreme configurations, $t=0$ and $t=1$. These cases represent, respectively, the situation of maximal overlap between the two classes and the scenario in which they are the most separated. Detailed specifications of the data-generation procedure, including the precise parameter values, are provided in the Supplementary Material. We again consider different sample sizes for each class, that is, $n_0=n_1\in\{250, 500,1000\}$.

Table \ref{tab:three-norm-all} shows the results for the three-dimensional setting, while Table \ref{tab:ten-norm-all} corresponds to the ten-dimensional case. Focusing on three dimensions, MST-RClass achieves better performance than both MST-Class and KNN, being surpassed only by LDA when $t=1$ and QDA. In ten dimensions, the situation changes: MST-Class outperforms MST-RClass for $t=0$, which is also reflected in the mean value of $h$ in this case, close to $9$, indicating that cross-validation correctly identifies the need to select larger subsamples. Additionally, KNN improves upon MST-Class and MST-RClass in $t=0$. Finally, there appears to be an inverse relationship between $h$ and $k$: larger $h$ corresponds to smaller $k$.

\begin{table}[!htbp]
	\centering
	\resizebox{\textwidth}{!}{%
	\begin{tabular}{c l c cc cc c c}
		\toprule
		Class size & $t$ & MST-Class & \multicolumn{2}{c}{MST-RClass} & \multicolumn{2}{c}{KNN} & LDA & QDA \\
		\cmidrule(lr){4-5} \cmidrule(lr){6-7}
		&  &  & Accuracy & $h$ & Accuracy & $k$ &  &  \\
		\midrule
		$n_0=n_1=250$ & 0 & 0.711 (0.039) & 0.754 (0.036) & 3.500 (3.010) & 0.747 (0.034) & 13.110 (5.393) & 0.492 (0.053) & 0.775 (0.036) \\
		& 1 & 0.997 (0.004) & 0.998 (0.004) & 1.830 (2.391) & 0.998 (0.003) & 15.150 (6.360) & 0.999 (0.003) & 0.998 (0.004) \\
		\midrule
		$n_0=n_1=500$ & 0 & 0.708 (0.026) & 0.762 (0.026) & 2.370 (2.223) & 0.752 (0.024) & 20.520 (6.779) & 0.500 (0.040) & 0.774 (0.024) \\
		& 1 & 0.997 (0.003) & 0.998 (0.003) & 2.140 (2.598) & 0.998 (0.003) & 17.630 (10.178) & 0.998 (0.003) & 0.998 (0.003) \\
		\midrule
		$n_0=n_1=1000$ & 0 & 0.705 (0.019) & 0.764 (0.018) & 2.333 (1.927) & 0.762 (0.020) & 29.242 (8.350) & 0.503 (0.032) & 0.777 (0.017) \\
		& 1 & 0.997 (0.003) & 0.998 (0.002) & 2.400 (2.871) & 0.998 (0.002) & 21.779 (14.816) & 0.998 (0.002) & 0.998 (0.002) \\
		\bottomrule
	\end{tabular}
}
\caption{Performance metrics for different methods across configurations of three-dimensional normal data ($d=3$), with varying number of observations per class.}
\label{tab:three-norm-all}
\end{table}

\begin{table}[!htbp]
	\centering
	\resizebox{\textwidth}{!}{%
		\begin{tabular}{c l c cc cc c c}
			\toprule
			Class size & $t$ & MST-Class & \multicolumn{2}{c}{MST-RClass} & \multicolumn{2}{c}{KNN} & LDA & QDA \\
			\cmidrule(lr){4-5} \cmidrule(lr){6-7}
			&  &  & Accuracy & $h$ & Accuracy & $k$ &  &  \\
			\midrule
			$n_0=n_1=250$ & 0 & 0.708 (0.041) & 0.690 (0.047) & 8.980 (2.108) & 0.866 (0.029) & 1.295 (0.707) & 0.499 (0.051) & 0.999 (0.003) \\
			& 1 & 1.000 (0.000) & 1.000 (0.000) & 1.000 (0.000) & 1.000 (0.000) & 19.000 (0.000) & 1.000 (0.000) & 1.000 (0.000) \\
			\midrule
			$n_0=n_1=500$ & 0 & 0.719 (0.031) & 0.702 (0.034) & 9.460 (1.594) & 0.891 (0.018) & 1.425 (0.847) & 0.500 (0.035) & 0.999 (0.002) \\
			& 1 & 1.000 (0.000) & 1.000 (0.000) & 1.000 (0.000) & 1.000 (0.000) & 27.000 (0.000) & 1.000 (0.000) & 1.000 (0.000) \\
			\midrule
			$n_0=n_1=1000$ & 0 & 0.734 (0.022) & 0.714 (0.024) & 9.900 (0.704) & 0.906 (0.012) & 1.400 (0.804) & 0.510 (0.023) & 0.999 (0.001) \\
			& 1 & 1.000 (0.000) & 1.000 (0.000) & 1.000 (0.000) & 1.000 (0.000) & 38.000 (0.000) & 1.000 (0.000) & 1.000 (0.000) \\
			\bottomrule
		\end{tabular}%
	}
	\caption{Performance metrics for different methods across configurations of ten-dimensional normal data ($d=10$), with varying number of observations per class.}
	\label{tab:ten-norm-all}
\end{table}

\subsubsection{Geometrically structured data}
Having assessed the behaviour of MST-Class and MST-RClass on normally distributed data, where no particular geometric structure might favour MST-based methods, we now turn to scenarios in which the classes lie on well-defined geometric objects, which might lend themselves better for MST-class underlying mechanism. These settings include intersecting planes, the intersection of a spiral surface with a plane, and toroidal structures. Such configurations provide a natural counterpart to the normal-data experiments, as they introduce class separation patterns driven by geometry rather than by a covariance structure. These settings are particularly suitable for studying the behaviour of MST-based methods, as they explicitly exploit the connectivity structure revealed by the MSTs, which can help separate the classes even in situations where standard classifiers face difficulties. We illustrate this through the experiments presented below.

\

\noindent\emph{Intersecting planes}

The following experiment is conducted on simulated data lying on two intersecting planes in $\mathbb{R}^3$. The first plane is always the $XY$-plane ($z=0$), where the $X$ and $Y$ coordinates are independently sample from the uniform distribution on $[-5,5]$. The second plane is generated similarly, then rotated by an angle 
$\theta$ about the $X$-axis and translated vertically by a height $ht$. Three configurations 
are considered:
\begin{itemize}[noitemsep]
    \item {Configuration 1:} $\theta =\pi/2$, $ht=0$.
    \item {Configuration 2:} $\theta=\pi/4$, $ht=1$.
    \item {Configuration 3:} $\theta=\pi/8$, $ht=1.5$.
\end{itemize}

For each configuration, we generate 100 datasets with $n_0=n_1=1000$ points on each plane and compare the performance of the proposed classifiers MST-Class and MST-RClass with the benchmark methods KNN and LDA. For each dataset, 70\% of the observations were randomly assigned to the training set, and the remaining 30\% were used for testing. The classification results are reported in Table~\ref{tab:planes}. Overall, MST-Class and KNN exhibit comparable accuracy across all configurations, with slightly higher values than those achieved by MST-RClass. Nevertheless, these differences remain small, suggesting that the robust subsampling strategy in MST-RClass does not lead to a substantial deterioration in classification performance.

\begin{table}[!htbp]
\centering
\resizebox{\textwidth}{!}{%
\begin{tabular}{c c cc cc c}
\toprule
Config. & MST-Class & \multicolumn{2}{c}{MST-RClass} & \multicolumn{2}{c}{KNN} & LDA \\
\cmidrule(lr){3-4} \cmidrule(lr){5-6}
 &  & Accuracy & $h$ & Accuracy & $k$ &  \\
\midrule
1 & 0.984 (0.006) & 0.982 (0.007) & 8.160 (2.943) & 0.984 (0.006) & 1.190 (0.692) & 0.500 (0.036) \\
2 & 0.977 (0.006) & 0.975 (0.007) & 8.200 (2.892) & 0.981 (0.006) & 1.200 (0.667) & 0.634 (0.018) \\
3 & 0.957 (0.010) & 0.953 (0.011) & 9.520 (1.560) & 0.976 (0.006) & 1.100 (0.414) & 0.877 (0.011) \\
\bottomrule
\end{tabular}
}
\caption{Performance metrics for different methods across three configurations of intersecting planes.}
\label{tab:planes}
\end{table}

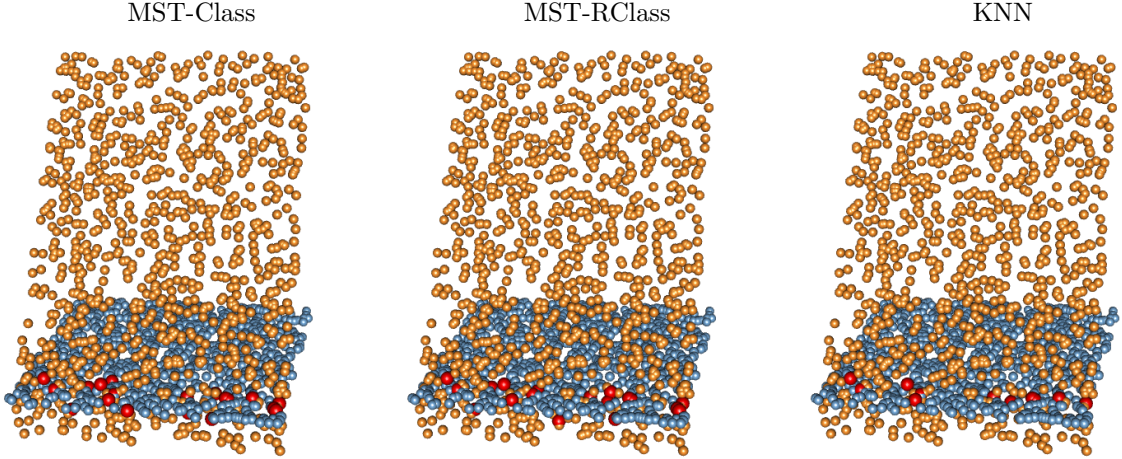
\begin{figure}[h]
	\centering
	\begin{tikzpicture}
		\node[anchor=south west,inner sep=0] (img) at (0,0)
		{\includegraphics[width=\textwidth,trim=0 160 0 110,
			clip]{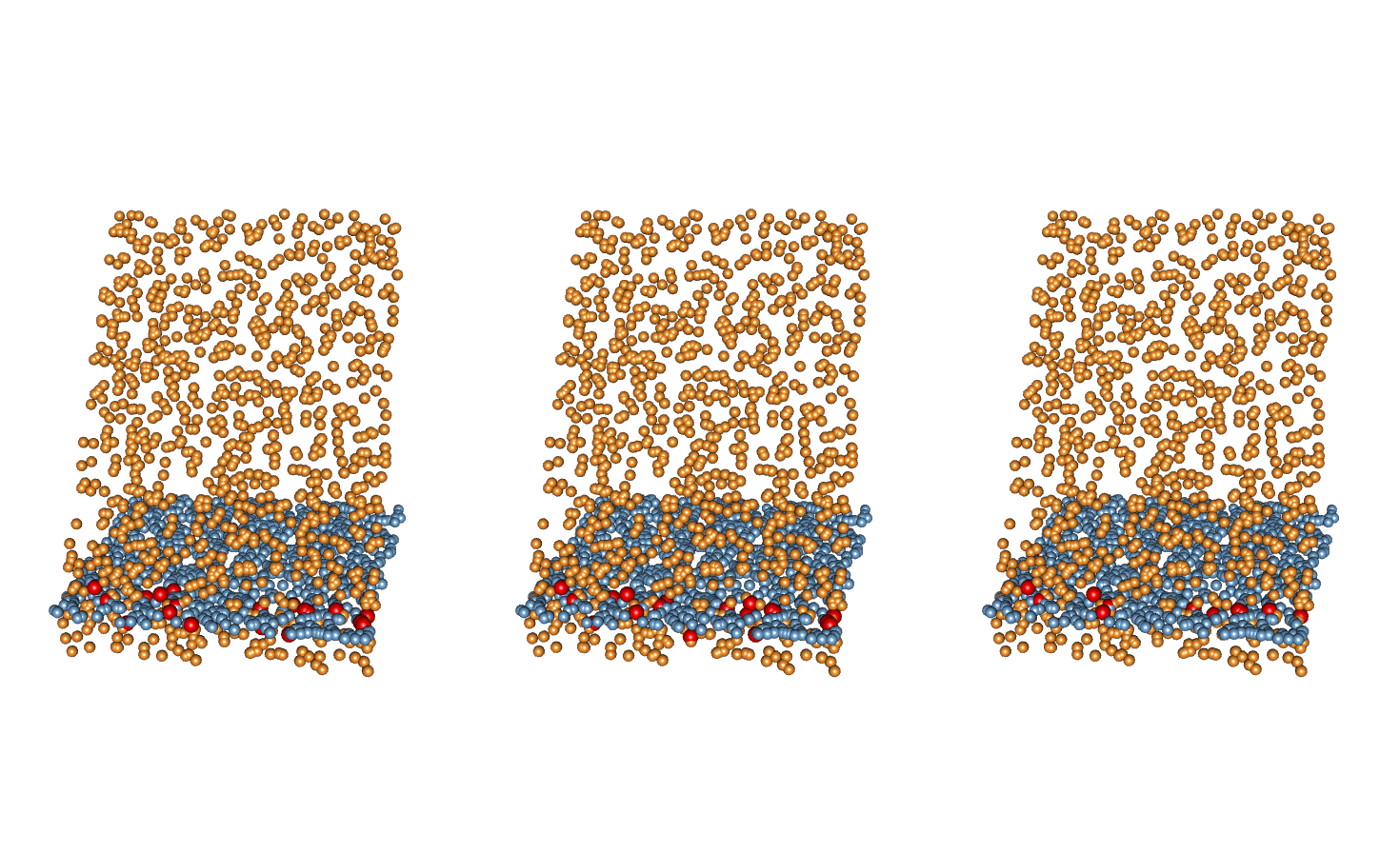}};
		
		\begin{scope}[x={(img.south east)},y={(img.north west)}]
			% títulos encima de cada gráfica
			\node at (0.19,0.9) {\footnotesize  MST-Class};
			\node at (0.525,0.9) {\footnotesize MST-RClass};
			\node at (0.86,0.9) {\footnotesize KNN};
		\end{scope}
	\end{tikzpicture}
	\caption{In red, misclassified points for a single representative dataset among the 100 generated for Configuration 1.}
	\label{fig:planes_all_configs}
\end{figure}

In Figure~\ref{fig:planes_all_configs}, we present the classification results for a single example dataset (out of the 100 generated) for Configuration 1. The figure displays the outcomes of each method (MST-Class, KNN, and MST-RClass), with misclassified points highlighted. As expected, the points that are challenging  for KNN largely coincide with those misclassified by MST-Class and MST-RClass and are primarily located in the intersecting regions of the planes. Additional figures are provided in the Supplementary Material.
%Resultados específicos para el dataset escogido de intersección de planos:
%Config1:0.965,0.966666666666667,0.97166666666666
%Config2: 0.97,0.966666666666667,0.975
%Config3:0.968333333333333,0.97,0.981666666666667

\

\noindent\emph{Intersecting spiral-plane}

Next, we evaluate MST-Class and MST-RClass on datasets lying on an intersecting spiral-plane structure, which present non-linear class separations. Three configurations are considered. In all cases, the first class consists of points lying on a two-dimensional plane, with $X$ coordinates sampled uniformly from $[-15,15]$ and  $Y$ coordinates from $[0,10]$. The second class consists of points arranged in a spiral plate, generated by varying the angular range $\phi$ and the radius parameter. The three specific configurations are:  
\begin{itemize}[noitemsep]
    \item  {Configuration 1:} $\phi = \pi$, radius $= 1.2$.
    \item  {Configuration 2:} $\phi = 1.5\pi$, radius $= 1$.
    \item  {Configuration 3:} $\phi = 2\pi$, radius $= 0.7$.
\end{itemize}

Again, for each configuration, we generate 100 datasets, with each class containing \(n_0 = n_1 = 1000\) points. Each dataset was randomly split into 70\% training and 30\% testing observations. The classification results are reported in Table~\ref{tab:spiral-plates}. Overall, MST-Class achieves slightly higher accuracy than MST-RClass across all configurations. Nevertheless, the differences are small, and MST-RClass remains competitive, exhibiting performance comparable to KNN while, as expected, LDA shows substantially lower accuracy in all configurations.

\begin{table}[!htbp]
\centering
\resizebox{\textwidth}{!}{%
\begin{tabular}{c c cc cc c}
\toprule
Config. & MST-Class & \multicolumn{2}{c}{MST-RClass} & \multicolumn{2}{c}{KNN} & LDA \\
\cmidrule(lr){3-4} \cmidrule(lr){5-6}
 &  & Accuracy & $h$ & Accuracy & $k$ &  \\
\midrule
1 & 0.979 (0.007) & 0.978 (0.007) & 8.390 (2.449) & 0.972 (0.007) & 1.220 (0.629) & 0.591 (0.066) \\
2 & 0.970 (0.008) & 0.968 (0.009) & 9.750 (1.095) & 0.968 (0.007) & 1.020 (0.200) & 0.564 (0.022) \\
3 & 0.955 (0.009) & 0.945 (0.017) & 7.400 (4.000) & 0.965 (0.007) & 1.100 (0.438) & 0.499 (0.044) \\
\bottomrule
\end{tabular}
}
\caption{Performance metrics for different methods across three configurations of intersecting spiral-plane}
\label{tab:spiral-plates}
\end{table}
%Resultados específicos para el dataset escogido de planchas en espiral:
%Config1:0.981666666666667,0.973333333333333,0.965
%Config2:0.96,0.956666666666667,0.953333333333333
%Config3:0.961666666666667,0.96,0.94666666666666

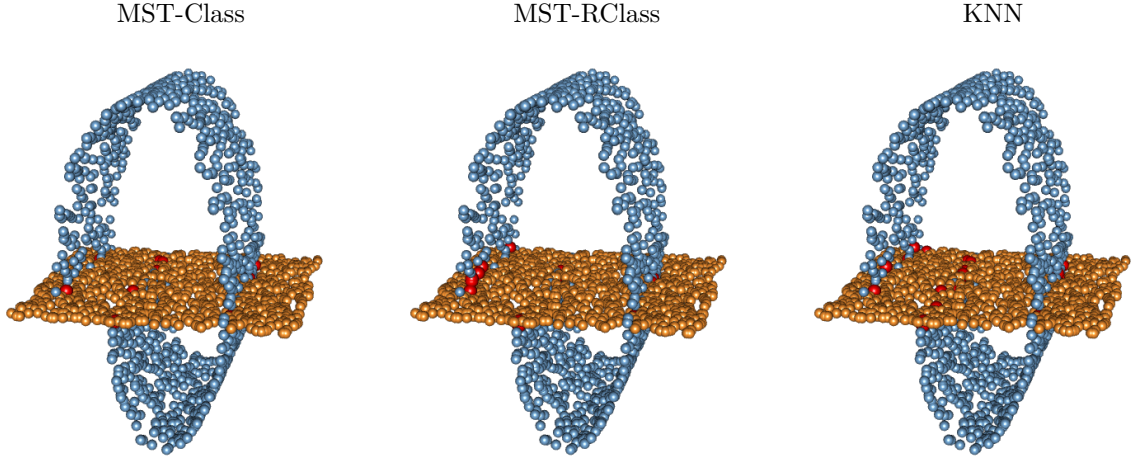
\begin{figure}[h]
	\centering
	\begin{tikzpicture}
		\node[anchor=south west,inner sep=0] (img) at (0,0)
		{\includegraphics[width=\textwidth,trim=0 160 0 100,
			clip]{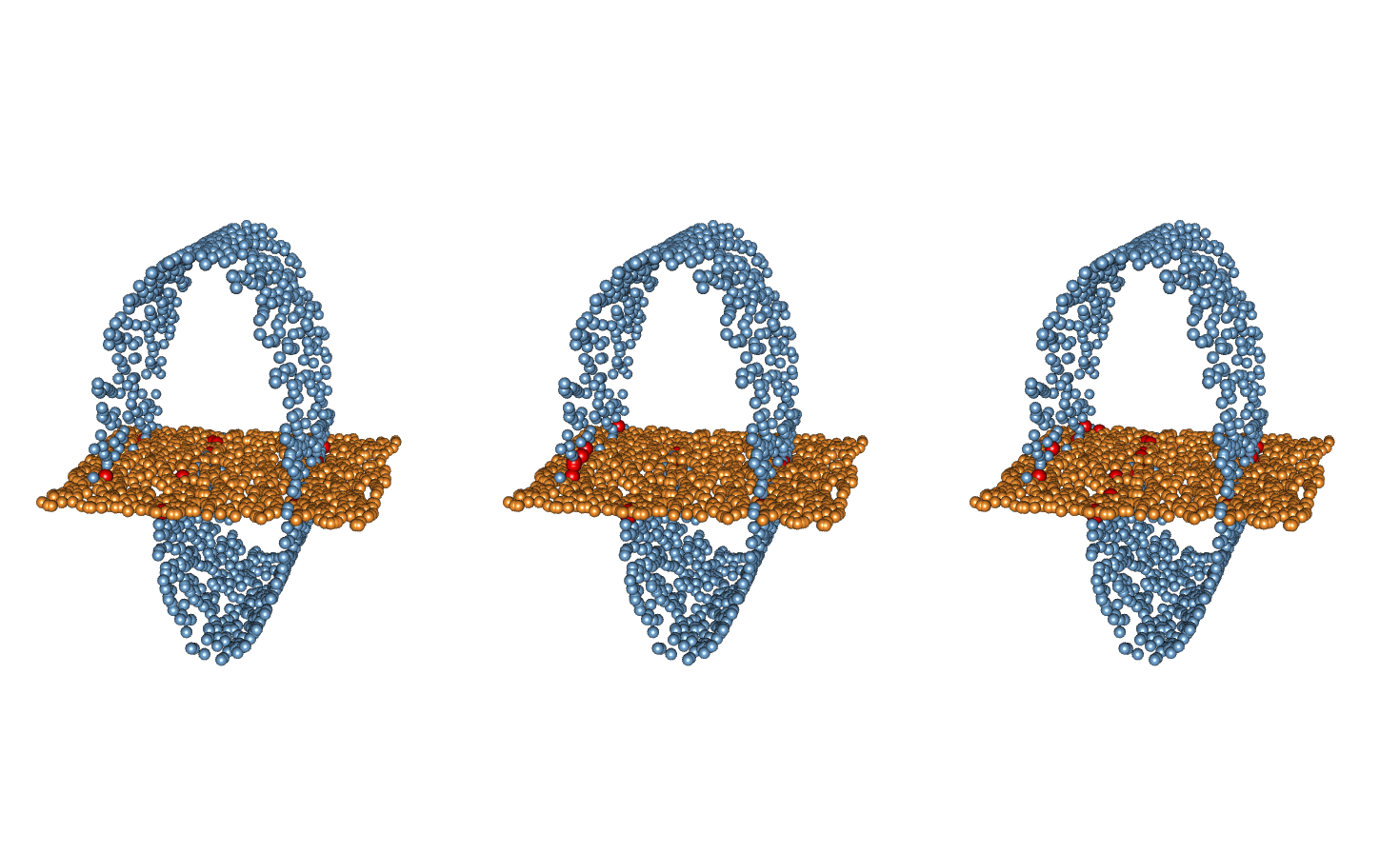}};
		
		\begin{scope}[x={(img.south east)},y={(img.north west)}]
			% títulos encima de cada gráfica
			\node at (0.17,0.9) {\footnotesize  MST-Class};
			\node at (0.505,0.9) {\footnotesize MST-RClass};
			\node at (0.84,0.9) {\footnotesize KNN};
		\end{scope}
	\end{tikzpicture}
	\caption{In red, misclassified points for a single representative dataset among the 100 generated for Configuration 1.}
	\label{fig:spiral_plate_all_configs}
\end{figure}

In Figure~\ref{fig:spiral_plate_all_configs}, we present the classification results for a single example dataset (out of the 100 generated) for Configuration 1. As in the intersecting planes, the points that are challenging for KNN largely coincide with those misclassified by MST-Class and MST-RClass, with most errors concentrated near the class boundaries. Additional figures are provided in the Supplementary Material.

For this geometric configuration we identified a different distribution of misclassified observations of MST-Class and MST-RClass with respect to KNN. Driven by this observation, we performed a complementary computational experiment on the spiral-plane datasets. Specifically, we selected one of the generated datasets for each configuration and split it into training and test sets 50 times, recording for each iteration the points misclassified by each method. 
The results, summarized in Table~\ref{tab:spiral_confusion}, reveal an interesting pattern: MST-Class and MST-RClass tend to make more errors in the plane, whereas KNN makes more errors along the spiral. This behavior is intriguing and suggests that each method exhibits complementary strengths and weaknesses depending on the class geometry.

\begin{table}[htbp]
	\centering
	\small % reduce el tamaño de la fuente
	\scalebox{0.8}{ % ajusta el factor según necesites
		\begin{tabular}{l c c c}
			\toprule
			Method & Configuration 1 & Configuration 2 & Configuration 3 \\
			\midrule
			MST-Class& 
			\begin{tabular}{c|cc}
				& \textbf{Pred 0} & \textbf{Pred 1} \\ \hline
				\textbf{True 0} & 964 & 36 \\
				\textbf{True 1} & 75 & 925
			\end{tabular} &
			\begin{tabular}{c|cc}
				& \textbf{Pred 0} & \textbf{Pred 1} \\ \hline
				\textbf{True 0} & 972 & 28 \\
				\textbf{True 1} & 135 & 865
			\end{tabular} &
			\begin{tabular}{c|cc}
				& \textbf{Pred 0} & \textbf{Pred 1} \\ \hline
				\textbf{True 0} & 995 & 5 \\
				\textbf{True 1} & 18 & 982
			\end{tabular} \\
			[3mm]
						MST-RClass &
			\begin{tabular}{c|cc}
				& \textbf{Pred 0} & \textbf{Pred 1} \\ \hline
				\textbf{True 0} & 950 & 50 \\
				\textbf{True 1} & 54 & 946
			\end{tabular} &
			\begin{tabular}{c|cc}
				& \textbf{Pred 0} & \textbf{Pred 1} \\ \hline
				\textbf{True 0} & 977 & 23 \\
				\textbf{True 1} & 136 & 864
			\end{tabular} &
			\begin{tabular}{c|cc}
				& \textbf{Pred 0} & \textbf{Pred 1} \\ \hline
				\textbf{True 0} & 997 & 3 \\
				\textbf{True 1} & 21 & 979
			\end{tabular} \\
			[3mm]
			
			KNN & 
			\begin{tabular}{c|cc}
				& \textbf{Pred 0} & \textbf{Pred 1} \\ \hline
				\textbf{True 0} & 931 & 69 \\
				\textbf{True 1} & 72 & 928
			\end{tabular} &
			\begin{tabular}{c|cc}
				& \textbf{Pred 0} & \textbf{Pred 1} \\ \hline
				\textbf{True 0} & 913 & 87 \\
				\textbf{True 1} & 62 & 938
			\end{tabular} &
			\begin{tabular}{c|cc}
				& \textbf{Pred 0} & \textbf{Pred 1} \\ \hline
				\textbf{True 0} & 970 & 30 \\
				\textbf{True 1} & 2 & 998
			\end{tabular} \\
			\bottomrule
		\end{tabular}
	}
	\caption{Confusion matrices for MST-Class, MST-RClass, and KNN on the spiral-plane datasets. Rows correspond to true classes and columns to predicted classes. Classes: 0 = spiral, 1 = plane.}
	\label{tab:spiral_confusion}
\end{table}

\

\noindent\emph{Intersecting tori}

As a final case within the simulated datasets, we assess MST-Class and MST-RClass on torus-shaped class structures. Three configurations of torus-shaped datasets are considered, with one or more torus per class. Each torus is defined by its minor radius $r$, major radius $R$, center coordinates $ct$ and optional rotation $rotx$ about the $X$-axis. The configurations are:

%\begin{itemize}[noitemsep]
%    \item {Configuration 1:} two torus; Torus 1: $r=0.5$, $R=2$; Torus 2: $r=0.5$, $R=2$, $ct=(2.8,0,0)$, $rotx=\pi/2$.
%    \item {Configuration 2:} three torus; Torus 1 (class 1): $r=0.5$, $R=2$; Torus 2 (class 2): $r=0.5$, $R=1.5$, $ct=(2.8,0,0)$, $rotx=\pi/2$; Torus 3 (class 2): $r=0.5$, $R=1.5$, $ct=(-2.8,0,0)$, $rotx=\pi/2$.
%    \item {Configuration 3:} four torus; Torus 1 (class 1): $r=0.5$, $R=2$; Torus 2 (class 2): $r=0.5$, $R=1.5$, $ct=(2.8,0,0)$, $rotx=\pi/2$; Torus 3 (class 2): $r=0.5$, $R=1.5$, $ct=(-1,0,-1)$, $rotx=\pi/2$; Torus 4 (class 2): $r=0.5$, $R=1.5$, $ct=(-1,0,1)$, $rotx=\pi/2$.
%\end{itemize}

\begin{itemize}[noitemsep]
    \item {Configuration 1 (two tori):} 
    Torus 1: $(r,R)=(0.5,2)$; 
    Torus 2: $(r,R)=(0.5,2)$, shifted by $ct=(2.8,0,0)$ and rotated by $rotx=\pi/2$.

    \item {Configuration 2 (three tori):} 
    Torus 1 (class 1): $(r,R)=(0.5,2)$; 
    Tori 2--3 (class 2): $(r,R)=(0.5,1.5)$, with 
    $ct=(\pm 2.8,0,0)$ and $rotx=\pi/2$.

    \item {Configuration 3 (four tori):} 
    Torus 1 (class 1): $(r,R)=(0.5,2)$; 
    Tori 2--4 (class 2): $(r,R)=(0.5,1.5)$, rotated by $rotx=\pi/2$, with centers
    $ct=(2.8,0,0)$, $(-1,0,-1)$, and $(-1,0,1)$.
\end{itemize}

For each configuration, we generate 100 datasets, with each class containing \(n_0 = n_1 = 1000\) points. Datasets were split randomly, with 70\% for training and 30\% for testing. Table \ref{tab:torus} presents the performance metrics for the three  configurations. Overall, MST-Class, MST-RClass and KNN achieve very high accuracy values, often reaching perfect classification. Both MST-Class and MST-RClass slightly improve over KNN in most configurations.

\begin{table}[!htbp]
\centering
\resizebox{\textwidth}{!}{%
\begin{tabular}{c c cc cc c}
\toprule
Config. & MST-Class & \multicolumn{2}{c}{MST-RClass} & \multicolumn{2}{c}{KNN} & LDA \\
\cmidrule(lr){3-4} \cmidrule(lr){5-6}
 &  & Accuracy & $h$ & Accuracy & $k$ &  \\
\midrule
1 & 1.000 (0.000) & 1.000 (0.000) & 1.000 (0.000) & 1.000 (0.001) & 30.460 (11.423) & 0.750 (0.019) \\
2 & 0.981 (0.006) & 0.984 (0.005) & 5.970 (3.421) & 0.978 (0.006) & 6.590 (3.947) & 0.496 (0.052) \\
3 & 0.971 (0.008) & 0.975 (0.008) & 7.560 (2.653) & 0.970 (0.007) & 4.640 (2.513) & 0.511 (0.020) \\
\bottomrule
\end{tabular}
}
\caption{Performance metrics for different methods across three configurations of torus.}
\label{tab:torus}
\end{table}

%Figures \ref{fig:torus_config2} and \ref{fig:torus_config3} show the misclassified points for each method. As in the previous structures, the errors mainly occur in the intersection regions between classes.

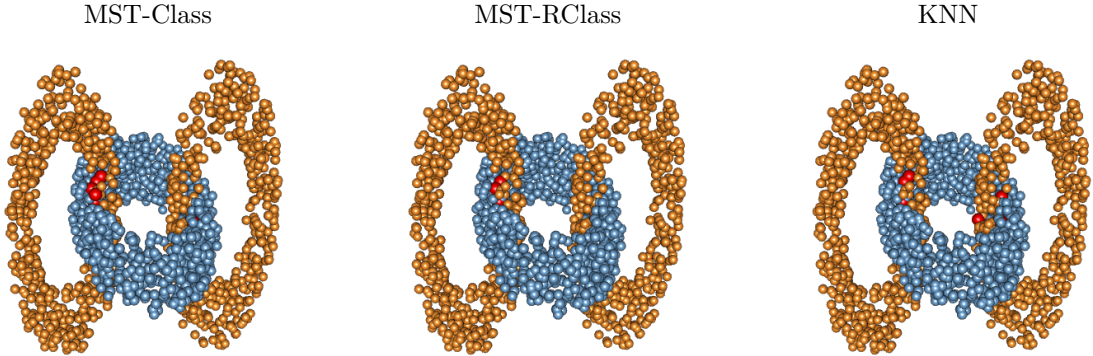
\begin{figure}[!htbp]
	\centering
	\begin{tikzpicture}

		\node[anchor=south west,inner sep=0] (img) at (0,0)
		{\includegraphics[
		width=\textwidth,
		trim=0 230 0 120,
		clip
		]{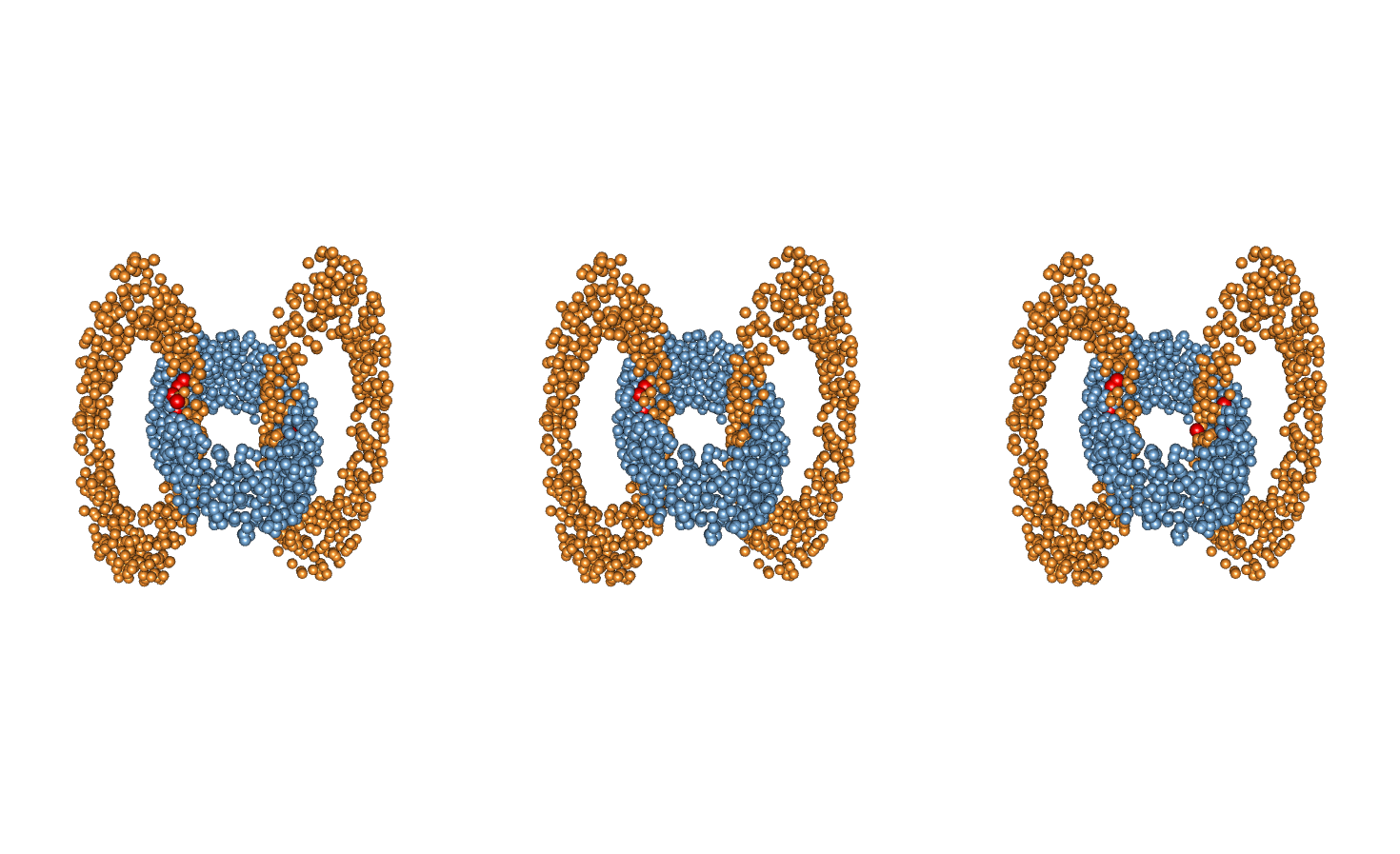}};
		
		\begin{scope}[x={(img.south east)},y={(img.north west)}]
			% títulos encima de cada gráfica
			\node at (0.17,0.85) {\footnotesize  MST-Class};
			\node at (0.50,0.85){\footnotesize MST-RClass}; 
			\node at (0.83,0.85) {\footnotesize KNN};
		\end{scope}
	\end{tikzpicture}
	\caption{In red, misclassified points for a single representative dataset among the 100 generated for Configuration 2.}
	\label{fig:torus_all_configs}
\end{figure}

In Figure~\ref{fig:torus_all_configs}, we present the classification results for a single example dataset (out of the 100 generated) for Configuration 2. Additional figures are provided in the Supplementary Material.
\vspace{-0.3cm}

Across all selected structured datasets (intersecting planes, intersecting spiral-plane and intersecting tori) MST-Class and MST-RClass generally achieve very high accuracy, often comparable to or slightly better than KNN. Considering all nine configurations, MST-based methods are outperformed by KNN in only three cases, with minor differences. These experiments on structured geometries confirm the initial motivation for using such datasets: the ability of MST-based methods to exploit the intrinsic geometric structure of the data through the connectivity captured by the MST.

Overall, these findings conclude the analysis on simulated data, demonstrating the robustness and competitiveness of the proposed MST-based classifiers across diverse settings.

\subsection{A case study: aircraft trajectories}\label{Section:trajectories}
To complement the insights gained from simulated experiments, we evaluate the performance of MST-Class and MST-RClass on a real-world dataset of aircraft trajectories. This case study illustrates how the MST-based approach can capture relationships among observations in a setting where the data exhibit an underlying geometric structure and contribute to the classification process.

The dataset considered is TrajAir \citep{trajectories}, which provides recorded trajectories of multiple aircraft operating at the Pittsburgh-Butler Regional Airport, located north of Pittsburgh, Pennsylvania. The dataset is organized into 3088 ``scenes'', each of which begins when at least one aircraft enters the detection range and ends when all aircraft have left the area or become inactive. Multiple aircraft identifiers may be present within a single scene, and the same aircraft can appear across different scenes.

To illustrate the application of MST-based classifiers on real data, we randomly select 100 individual trajectories from the dataset, each containing approximately 500 points, in order to keep computation times manageable. In our setting, the explanatory variables consist solely of the geographic coordinates of the aircraft positions. From these trajectories, we construct pairs of trajectories, where each pair consists of two aircraft paths and is treated as a binary classification problem: each point is labeled according to the aircraft it belongs to.

The analysis is divided into two parts. In the first part, we randomly form 30 trajectory pairs, providing a general comparison of the considered classification methods (MST-Class, MST-RClass, and KNN) under typical conditions. In the second part, we select an additional set of 30 trajectory pairs for which standard approaches such as KNN perform poorly. This more challenging subset allows us to assess whether MST-based classifiers can offer improvements in difficult scenarios where classical methods struggle. For each trajectory pair, a training/test split is performed 10 times (70\% training, 30\% test), and the reported results correspond to the average accuracy of each method across these 10 partitions, with the standard deviation shown in parentheses. 
Although this experimental setup does not correspond to an actual operational task, it provides a clear and interpretable framework for evaluating the behavior of MST-based classifiers on real trajectory data.

The results for the first part of the analysis are reported in the Supplementary Material. In this relatively easy scenario, MST-Class matches or outperforms KNN in 24 cases, while MST-RClass does so in 16 cases. Overall, for these relatively easy cases, the average accuracies are 0.9969 for MST-Class, 0.9944 for MST-RClass, and 0.9930 for KNN, confirming that all methods perform very well in this scenario.  Analyzing the behavior of MST-RClass specifically, we observe that the parameter $h$ attains a mean value of 8.4633, indicating that cross-validation correctly identifies the need for relatively large subsamples to capture the global structure of the trajectories without overfitting to local noise.

\begin{figure}[!htbp]
	\centering
		
	% --------- Subfigure 1 ---------
	\begin{subfigure}[b]{0.7\textwidth}
		\centering
		\begin{tikzpicture}
			% Imagen
			\node[anchor=south west,inner sep=0] (img) at (0,0)
			{\includegraphics[width=\textwidth]{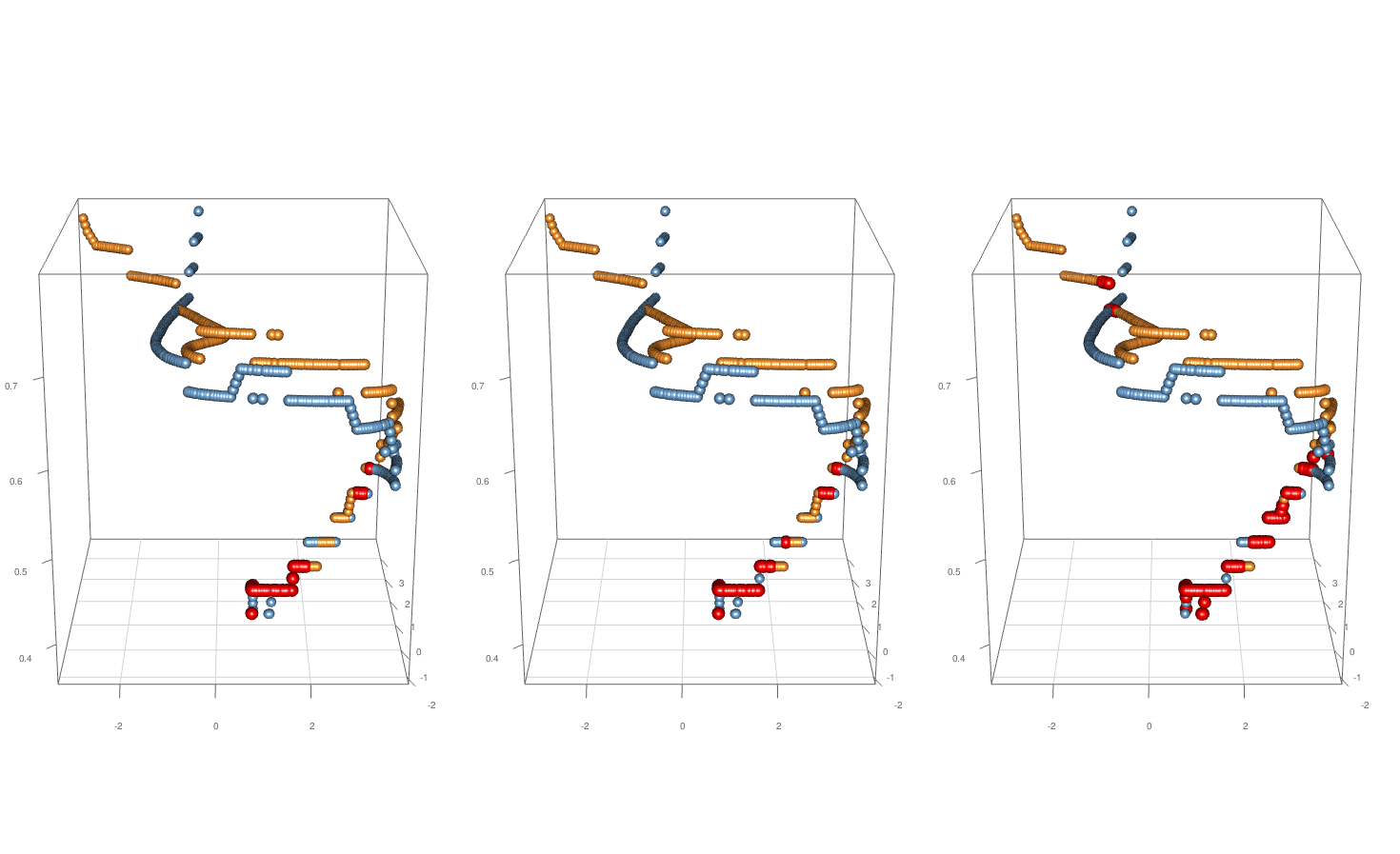}};
			
			% Títulos MST-Class, KNN, MST-RClass sobre la imagen
			\begin{scope}[x={(img.south east)},y={(img.north west)}, every node/.style={font=\footnotesize, text centered}]
				\node at (0.17,0.85) {MST-Class};
				\node at (0.5,0.85) {MST-RClass};
				\node at (0.83,0.85) {KNN};
				% Título principal Trajectory pair X
				\node at (0.5,0.92) {Trajectory pair 1};
			\end{scope}
		\end{tikzpicture}
	\end{subfigure}
	\vspace{-0.9cm}
	
	% --------- Subfigure 2 ---------
	\begin{subfigure}[b]{0.7\textwidth}
		\centering
		\begin{tikzpicture}
			\node[anchor=south west,inner sep=0] (img) at (0,0)
			{\includegraphics[width=\textwidth]{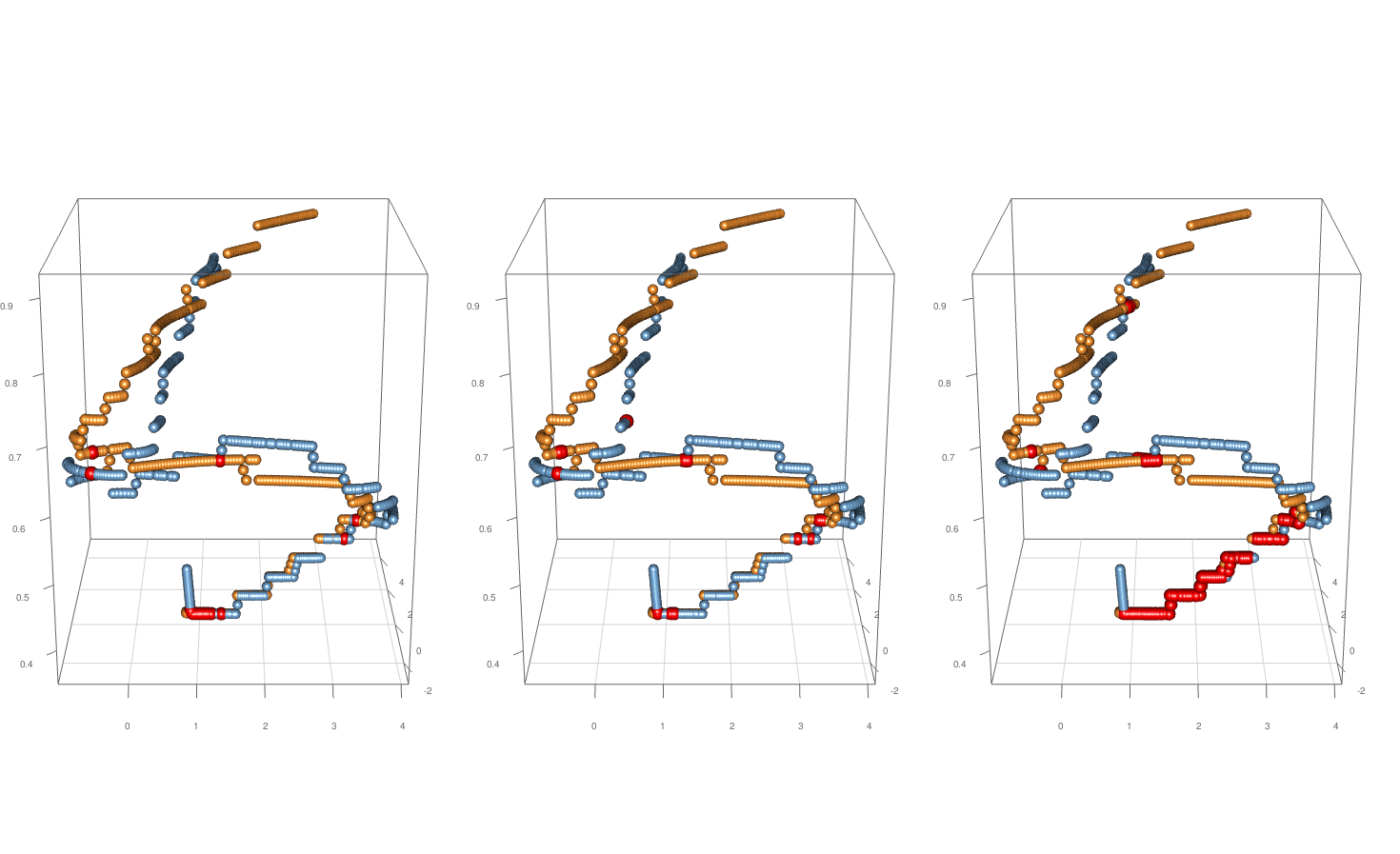}};
			\begin{scope}[x={(img.south east)},y={(img.north west)}, every node/.style={font=\footnotesize, text centered}]
				\node at (0.17,0.85) {MST-Class};
\node at (0.5,0.85) {MST-RClass};
\node at (0.83,0.85) {KNN};
				\node at (0.5,0.92) {Trajectory pair 9};
			\end{scope}
		\end{tikzpicture}
	\end{subfigure}
		\vspace{-0.9cm}
		
	% --------- Subfigure 3 ---------
	\begin{subfigure}[b]{0.7\textwidth}
		\centering
		\begin{tikzpicture}
			\node[anchor=south west,inner sep=0] (img) at (0,0)
			{\includegraphics[width=\textwidth]{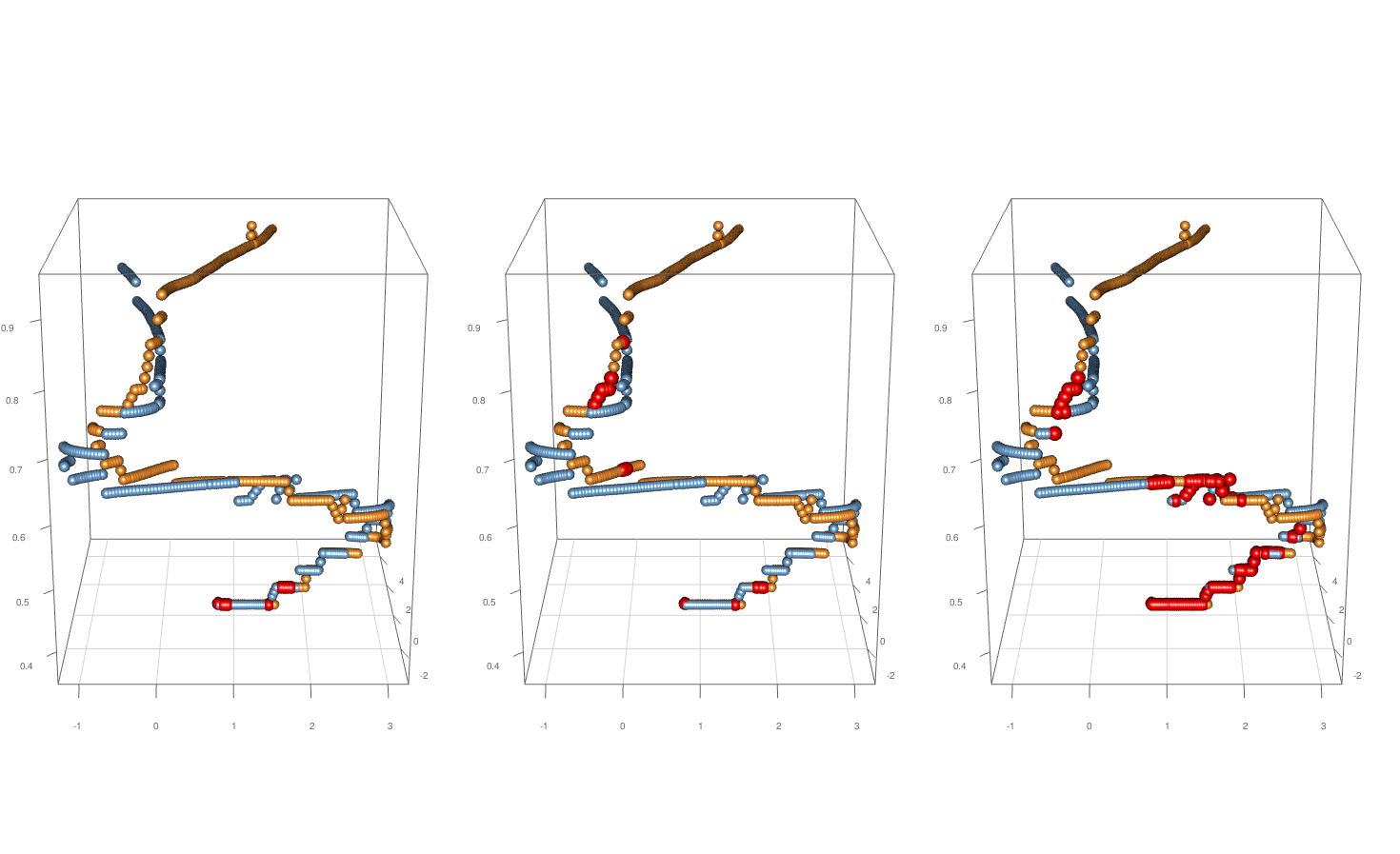}};
			\begin{scope}[x={(img.south east)},y={(img.north west)}, every node/.style={font=\footnotesize, text centered}]
				\node at (0.17,0.85) {MST-Class};
\node at (0.5,0.85) {MST-RClass};
\node at (0.83,0.85) {KNN};
				\node at (0.5,0.92) {Trajectory pair 10};
			\end{scope}
		\end{tikzpicture}
	\end{subfigure}
		\vspace{-0.9cm}
		
		% --------- Subfigure 4 ---------
			\begin{subfigure}[b]{0.7\textwidth}
			\centering
			\begin{tikzpicture}
				\node[anchor=south west,inner sep=0] (img) at (0,0)
				{\includegraphics[width=\textwidth]{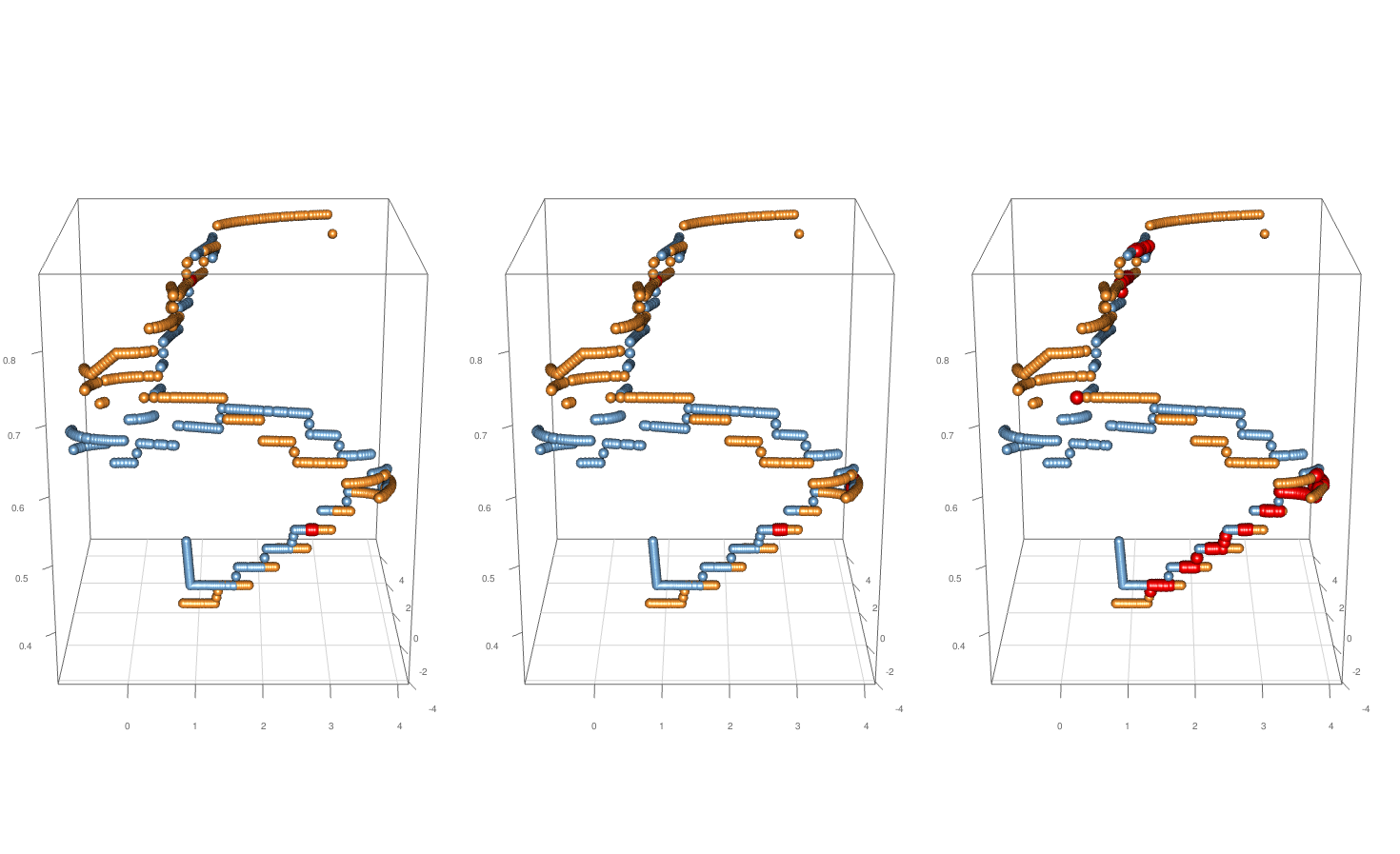}};
				\begin{scope}[x={(img.south east)},y={(img.north west)}, every node/.style={font=\footnotesize, text centered}]
				\node at (0.17,0.85) {MST-Class};
\node at (0.5,0.85) {MST-RClass};
\node at (0.83,0.85) {KNN};
					\node at (0.5,0.92) {Trajectory pair 17};
				\end{scope}
			\end{tikzpicture}
		\end{subfigure}
	\vspace{-1cm}
	\caption{Visualization of four trajectory pairs selected from the challenging set (i.e., pairs for which KNN performed worst). 
		For each pair, misclassified points are shown in red for each method.}
	\label{fig:trajectories_all}
\end{figure}

For the second part, we focus on most challenging cases. Specifically, we select the 30 trajectory pairs for which KNN produced the lowest accuracies. Table~\ref{tab:aircraft-worst} presents the results. Accuracies of MST-Class and MST-RClass that are equal to or higher than the corresponding KNN accuracy for the same trajectory pair are highlighted in bold. Here, MST-Class outperforms KNN in all 30 cases, while MST-RClass does so in 29 cases. The differences are substantial, contrasting with the minor variations observed in the random sample; for instance, in the fourth trajectory pair, accuracies of 0.997 with MST-Class and 0.995 with MST-RClass drop to 0.904 with KNN. Overall, the average accuracies are 0.9476 for MST-Class, 0.9493 for MST-RClass, and 0.8944 for KNN, confirming the potential of MST-based classifiers to exploit the underlying structure and overcome the difficulties faced by KNN in these cases. Getting into the specific behavior of MST-RClass we see that, in most cases, the parameter $h$ attains a mean of 10, reflecting the increased need for large subsamples that allow the MST to capture a more global view of the trajectory structure and avoid overly local or ``myopic'' classifications. This effect is also clearly visible in the selected trajectory pairs shown in Figure~\ref{fig:trajectories_all}, where misclassified points for each method are highligthed.

\begin{table}[!htbp] \centering \resizebox{\textwidth}{!}{%
		\begin{tabular}{c c cc cc} \toprule 
			Pair & MST-Class & \multicolumn{2}{c}{MST-RClass} & \multicolumn{2}{c}{KNN} \\ \cmidrule(lr){3-4} \cmidrule(lr){5-6}
			& & Accuracy & $h$ & Accuracy & $k$ \\ \midrule 
			1 & \textbf{0.894 (0.014)} & \textbf{0.941 (0.016)} & 10.000 (0.000) & 0.805 (0.022) & 4.400 (2.634) \\
			2 & \textbf{0.922 (0.014)} & \textbf{0.910 (0.017)} & 8.500 (2.415) & 0.864 (0.013) & 2.500 (0.527) \\
			3 & \textbf{0.956 (0.009)} & \textbf{0.966 (0.011)} & 10.000 (0.000) & 0.872 (0.017) & 5.000 (2.667) \\
			4 & \textbf{0.922 (0.020)} & \textbf{0.934 (0.014)} & 9.500 (1.581) & 0.880 (0.020) & 2.600 (0.966) \\
			5& \textbf{0.917 (0.014)} & \textbf{0.912 (0.016)} & 10.000 (0.000) & 0.880 (0.018) & 2.200 (0.789) \\
			6 & \textbf{0.947 (0.011)} & \textbf{0.938 (0.012)} & 10.000 (0.000) & 0.885 (0.018) & 2.300 (0.483) \\
			7 & \textbf{0.951 (0.010)} & \textbf{0.927 (0.015)} & 8.000 (2.582) & 0.889 (0.014) & 1.700 (0.949) \\
			8 & \textbf{0.947 (0.015)} & \textbf{0.958 (0.007)} & 10.000 (0.000) & 0.889 (0.014) & 7.800 (3.393) \\
			9 & \textbf{0.985 (0.007)} & \textbf{0.985 (0.008)} & 10.000 (0.000) & 0.890 (0.020) & 2.000 (0.000) \\
			10 & \textbf{0.991 (0.006)} & \textbf{0.992 (0.009)} & 10.000 (0.000) & 0.890 (0.022) & 1.800 (0.633) \\
			11 & \textbf{0.942 (0.010)} & \textbf{0.912 (0.033)} & 7.900 (3.479) & 0.890 (0.019) & 2.600 (0.516) \\
			12 & \textbf{0.957 (0.008)} & \textbf{0.954 (0.011)} & 10.000 (0.000) & 0.891 (0.017) & 2.400 (1.265) \\
			13 & \textbf{0.959 (0.012)} & \textbf{0.958 (0.007)} & 10.000 (0.000) & 0.897 (0.018) & 2.000 (0.000) \\
			14 & \textbf{0.947 (0.021)} & \textbf{0.955 (0.019)} & 9.000 (2.108) & 0.898 (0.024) & 2.700 (0.675) \\
			15 & \textbf{0.963 (0.015)} & \textbf{0.968 (0.012)} & 10.000 (0.000) & 0.899 (0.021) & 2.300 (0.483) \\
			16 & \textbf{0.945 (0.013)} & \textbf{0.939 (0.018)} & 10.000 (0.000) & 0.900 (0.020) & 3.600 (1.838) \\
			17& \textbf{0.997 (0.003)} & \textbf{0.995 (0.004)} & 10.000 (0.000) & 0.904 (0.032) & 2.500 (0.972) \\
			18 & \textbf{0.899 (0.023)} & \textbf{0.931 (0.013)} & 10.000 (0.000) & 0.904 (0.016) & 2.000 (0.667) \\
			19 & \textbf{0.981 (0.010)} & \textbf{0.982 (0.010)} & 10.000 (0.000) & 0.904 (0.017) & 2.600 (0.699) \\
			20 & \textbf{0.947 (0.015)} & \textbf{0.950 (0.012)} & 10.000 (0.000) & 0.905 (0.020) & 3.300 (1.252) \\
			21 & \textbf{0.934 (0.018)} & \textbf{0.925 (0.018)} & 10.000 (0.000) & 0.906 (0.014) & 1.800 (0.789) \\
			22 & \textbf{0.934 (0.025)} & \textbf{0.955 (0.015)} & 10.000 (0.000) & 0.908 (0.014) & 2.300 (0.483) \\
			23 & \textbf{0.958 (0.014)} & \textbf{0.951 (0.009)} & 9.500 (1.581) & 0.909 (0.014) & 2.600 (1.075) \\
			24 & \textbf{0.982 (0.008)} & \textbf{0.977 (0.016)} & 10.000 (0.000) & 0.910 (0.014) & 1.600 (0.699) \\
			25 & \textbf{0.928 (0.006)} & \textbf{0.951 (0.006)} & 10.000 (0.000) & 0.910 (0.014) & 4.200 (2.150) \\
			26 & \textbf{0.922 (0.016)} & 0.904 (0.038) & 9.000 (2.108) & 0.910 (0.017) & 2.600 (0.516) \\
			27 & \textbf{0.945 (0.005)} & \textbf{0.944 (0.005)} & 10.000 (0.000) & 0.910 (0.019) & 2.500 (0.707) \\
			28 & \textbf{0.927 (0.017)} & \textbf{0.925 (0.014)} & 10.000 (0.000) & 0.910 (0.016) & 3.800 (1.874) \\
			29 & \textbf{0.976 (0.008)} & \textbf{0.978 (0.008)} & 10.000 (0.000) & 0.911 (0.022) & 2.000 (0.667) \\
			30 & \textbf{0.953 (0.010)} & \textbf{0.963 (0.015)} & 9.500 (1.581) & 0.912 (0.017) & 1.800 (0.422) \\ \bottomrule 
	\end{tabular} }
	\caption{Classification results for MST-Class, MST-RClass, and KNN across the 30 trajectory pairs where KNN performed worst, sorted from lowest to highest KNN accuracy.} \label{tab:aircraft-worst}
\end{table}

\section{Conclusions and future research}\label{Section:conclusions}

In this paper, we introduce two new classifiers based on MST: MST-Class and MST-RClass. These methods exploit the connectivity and global structure of the data. We evaluated these methods on both simulated datasets and on a real-world dataset of aircraft trajectories.

The computational study on simulated data showed that MST-based classifiers are competitive, often matching or slightly outperforming traditional methods like KNN, especially when the intrinsic geometric structure plays a significant role. Additionally, MST-RClass offers increased robustness to mislabeled points while also improving computational efficiency compared to MST-Class.

Analysis of aircraft trajectory data further demonstrated the practical advantages of MST-based methods. Both MST-Class and MST-RClass outperformed KNN in challenging segments, such as near the landing zone, highlighting the advantage of considering the trajectory graph as a whole, rather than relying solely on local neighbors. Overall, these results suggest that MST-based classifiers are especially well-suited for datasets with complex geometric or network-like structures. They offer a robust and effective alternative to standard distance-based methods.

While the empirical results for MST-Class and MST-RClass are promising, formally establishing their asymptotic properties remains a challenging direction for future research. The primary hurdle lies in the underlying combinatorial structure of Minimum Spanning Trees. To explore this, we conducted preliminary numerical analyses aimed at identifying dominant structural patterns in large graphs to better understand the conformity shifts that occur when a new observation is added. Our tests revealed persistent structural complexities that resist simple asymptotic characterization. For instance, the MST node degree distribution, while stabilizing for large graphs, remains heterogeneous. They settle at roughly 50\% for degree 1, splitting the remainder evenly between degrees 2 and 3, and retaining a non-vanishing proportion of degree-4 nodes. Furthermore, when analyzing how an MST updates upon receiving a new observation, we found that most updates are simple: either adding a single edge to the nearest neighbor or ``splitting'' an existing edge. Yet, there remains a significant, non-vanishing probability of more complex topological rearrangements. Ultimately, these persistent combinatorial variations complicate the derivation of theoretical guarantees for MST-based classifiers.

\section*{Acknowledgments}
Projects PID2020-116587GB-I00, PID2024-158017NB-I00 and PID2021-124030NB-C32 funded by MICIU/AEI/10.13039/501100011033 and by FEDER, UE. This research was also funded by Grupos de Referencia Competitiva ED431C-2021/24 and ED431C-2025/03 from the Consellería de Cultura, Educación e Universidades, Xunta de Galicia. Iria Rodríguez-Acevedo acknowledges the support from Consellería de Cultura, Educación, Formación Profesional e Universidades, Xunta de Galicia, through grant ED481A-2023-061. 

ChatGPT (OpenAI, GPT-4o) and DeepL were used during the preparation of this manuscript for the sole purpose of language improvement. 

\paragraph{Conflict of interest} The authors declare no competing interests.

%\appendix
%\section{Example Appendix Section}
%\label{app1}


%%       \cite{<label>} ==> [1]

%%
%Example citation, See \cite{lamport94}.



\bibliographystyle{elsarticle-harv} 
\bibliography{biblio} 

@article{gan21,
author = {Ghurumuruhan Ganesan},
title = {{Minimum spanning trees of random geometric graphs with location dependent weights}},
volume = {27},
journal = {Bernoulli},
number = {4},
publisher = {Bernoulli Society for Mathematical Statistics and Probability},
pages = {2473 -- 2493},
year = {2021},
doi = {10.3150/20-BEJ1318},
URL = {https://doi.org/10.3150/20-BEJ1318}
}

@ARTICLE{gra85,
  author={Graham, R.L. and Hell, Pavol},
  journal={Annals of the History of Computing}, 
  title={On the History of the Minimum Spanning Tree Problem}, 
  year={1985},
  volume={7},
  number={1},
  pages={43-57},
  doi={10.1109/MAHC.1985.10011}}

@book{yuk98,
  author    = {Yukich, Joseph E.},
  title     = {Probability Theory of Classical Euclidean Optimization Problems},
  year      = {1998},
  publisher = {Springer},
  address   = {Berlin},
  series    = {Lecture Notes in Mathematics},
  volume    = {1675},
  isbn      = {978-3-540-63666-3},
  url       = {link.springer.com}
}

@article{lee00,
title = {Rate of convergence of power-weighted Euclidean minimal spanning trees},
journal = {Stochastic Processes and their Applications},
volume = {86},
number = {1},
pages = {163-176},
year = {2000},
issn = {0304-4149},
doi = {https://doi.org/10.1016/S0304-4149(99)00091-5},
url = {https://www.sciencedirect.com/science/article/pii/S0304414999000915},
author = {Sungchul Lee}
}

@article{yuk00,
title = {Asymptotics for weighted minimal spanning trees on random points},
journal = {Stochastic Processes and their Applications},
volume = {85},
number = {1},
pages = {123-138},
year = {2000},
issn = {0304-4149},
doi = {https://doi.org/10.1016/S0304-4149(99)00068-X},
url = {https://www.sciencedirect.com/science/article/pii/S030441499900068X},
author = {J.E. Yukich}
}

@article{pen03,
author = {Mathew D. Penrose and J. E. Yukich},
title = {{Weak laws of large numbers in geometric probability}},
volume = {13},
journal = {The Annals of Applied Probability},
number = {1},
publisher = {Institute of Mathematical Statistics},
pages = {277 -- 303},
year = {2003},
doi = {10.1214/aoap/1042765669},
URL = {https://doi.org/10.1214/aoap/1042765669}
}

@article{prim,
  title={Shortest connection networks and some generalizations},
  author={Prim, Robert Clay},
  journal={The Bell System Technical Journal},
  volume={36},
  number={6},
  pages={1389--1401},
  year={1957},
  publisher={Nokia Bell Labs}
}

@article{kruskal,
  title={On the shortest spanning subtree of a graph and the traveling salesman problem},
  author={Kruskal, Joseph B},
  journal={Proceedings of the American Mathematical society},
  volume={7},
  number={1},
  pages={48--50},
  year={1956}
}

@article{caret,
  title={Package ‘caret’},
  author={Kuhn, Max and Wing, Jed and Weston, Steve and Williams, Andre and Keefer, Chris and Engelhardt, Allan and Cooper, Tony and Mayer, Zachary and Kenkel, Brenton and Team, R Core and others},
  journal={The R Journal},
  volume={223},
  number={7},
  year={2020}
}

@article{bazlamaccci,
  title={Minimum-weight spanning tree algorithms a survey and empirical study},
  author={Bazlama{\c{c}}c{\i}, C{\"u}neyt F and Hindi, Khalil S},
  journal={Computers \& Operations Research},
  volume={28},
  number={8},
  pages={767--785},
  year={2001},
  publisher={Elsevier}
}

@ARTICLE{zah71,
  author={Zahn, C.T.},
  journal={IEEE Transactions on Computers}, 
  title={Graph-Theoretical Methods for Detecting and Describing Gestalt Clusters}, 
  year={1971},
  volume={C-20},
  number={1},
  pages={68-86},
  doi={10.1109/T-C.1971.223083}}

@article{xu01,
  title={Minimum Spanning Trees for Gene Expression Data Clustering},
  author={Ying Xu and Victor Olman and Dong Xu},
  journal={Genome Informatics},
  volume={12},
  number={ },
  pages={24-33},
  year={2001},
  doi={10.11234/gi1990.12.24}
}

@article{sa17,
title = {Sequential image segmentation based on minimum spanning tree representation},
journal = {Pattern Recognition Letters},
volume = {87},
pages = {155-162},
year = {2017},
note = {Advances in Graph-based Pattern Recognition},
issn = {0167-8655},
doi = {https://doi.org/10.1016/j.patrec.2016.06.001},
url = {https://www.sciencedirect.com/science/article/pii/S0167865516301192},
author = {Ali Saglam and Nurdan Akhan Baykan}
}

@article{li20,
  author    = {Li, J. and Wang, X. and Wang, X.},
  title     = {A scaled-MST-based clustering algorithm and application on image segmentation},
  journal   = {J. Intell. Inf. Syst.},
  volume    = {54},
  pages     = {501--525},
  year      = {2020},
  doi       = {10.1007/s10844-019-00572-x},
  url       = {https://doi.org/10.1007/s10844-019-00572-x}
}

@article{gag25,
  author    = {Gagolewski, M. and Cena, A. and Bartoszuk, M. and others},
  title     = {Clustering with Minimum Spanning Trees: How Good Can It Be?},
  journal   = {J. Classif.},
  volume    = {42},
  pages     = {90--112},
  year      = {2025},
  doi       = {10.1007/s00357-024-09483-1},
  url       = {https://doi.org/10.1007/s00357-024-09483-1}
}

@book{knn,
  title={Discriminatory analysis: nonparametric discrimination, consistency properties},
  author={Fix, Evelyn},
  volume={1},
  year={1985},
  publisher={USAF school of Aviation Medicine}
}

@inproceedings{mst1,
  title={Complex network-based data classification using minimum spanning tree metric and optimization},
  author={Saire, Josimar Chire and Zhao, Liang},
  booktitle={2023 International Joint Conference on Neural Networks (IJCNN)},
  pages={1--7},
  year={2023},
  organization={IEEE}
}

@article{mst2,
  title={Complex Networks for Pattern-Based Data Classification},
  author={Chire, Josimar and Mahmood, Khalid and Liang, Zhao},
  journal={arXiv preprint arXiv:2503.05772},
  year={2025}
}

@Manual{R,
    title = {R: A Language and Environment for Statistical Computing},
    author = {{R Core Team}},
    organization = {R Foundation for Statistical Computing},
    address = {Vienna, Austria},
    year = {2024},
    url = {https://www.R-project.org/},
  }

@Manual{igraph,
  title = {{igraph}: Network Analysis and Visualization in R},
  author = {Gábor Csárdi and Tamás Nepusz and Vincent Traag and Szabolcs Horvát and Fabio Zanini and Daniel Noom and Kirill Müller},
  year = {2025},
  note = {R package version 2.1.1},
  doi = {10.5281/zenodo.7682609},
  url = {https://CRAN.R-project.org/package=igraph},
}

@misc{trajectories,
  title={Trajair: A general aviation trajectory dataset},
  author={Patrikar, Jay and Moon, Brady and Ghosh, Sourish and Oh, Jean and Scherer, Sebastian},
  year={2021},
  publisher={Jun}
}

\end{document}

\begin{frontmatter}
		\title{Supplementary Material: A new classification method based on Minimum Spanning Trees}
		\author[1,2]{Julio González-Díaz}
		\author[1,2]{Beatriz Pateiro-López}
		\author[1,2]{Iria Rodríguez-Acevedo}
		\address[1]{Department of Statistics, Mathematical Analysis and Optimization and MODESTYA Research Group, University of Santiago de Compostela, Galicia, Spain}
		\address[2]{CITMAga (Galician Center for Mathematical Research and Technology), Santiago de Compostela, Spain}
	\end{frontmatter}
	
	\beginsupplementary

This Supplementary Material presents the technical specifications of the multivariate normal data generation process employed in the simulation study, providing details that were omitted from the main text. It also includes extended visualizations of results for the geometrically structured datasets (intersecting planes, spiral plates, and tori) and performance metrics for the aircraft trajectory case study.
	
Recall that the computational study is divided into two main parts. In the first part, we conduct a comprehensive comparison between MST-Class and MST-RClass, evaluating their performance both relative to each other and against several widely used classification techniques on simulated datasets.
In the second part, we present a case study based on real aircraft trajectory data, where the underlying data distribution is unknown and potentially more challenging.

In the first part of the study, two main types of scenarios are considered. The first consists of multivariate normal data, which follow a standard statistical structure and do not provide any particular advantage to MST-based classifiers. The second involves datasets with a strong underlying geometric organization, where MST-based classifiers may be more competitive than standard methods.

The Supplementary Material is organized as follows:

\begin{itemize}[leftmargin=1.5em]
	\item \textbf{Section S1: Normal data generation}
	\begin{itemize}
		\item Two dimensions
		\item Three dimensions
		\item Ten dimensions
	\end{itemize}
	\item \textbf{Section S2: Figures for geometrically structured data}
			\begin{itemize}
		\item Intersecting planes
		\item Intersecting spiral-plane
		\item Intersecting tori
	\end{itemize}
	
	\item \textbf{Section S3: Aircraft trajectory metrics}
\end{itemize}

%Presentar lo que viene a continuación y quizás antes poner el esquema del estudio computacional? Puede quedar bien creo.

\section{Normal data generation}\label{Section:normal-data}

In this section, we focus on the parameters of the multivariate normal data employed in the simulation study. Specifically, we consider a binary classification scenario, where each class follows a pre-specified multivariate normal distribution. Three scenarios are considered, corresponding to feature spaces of two, three, and ten dimensions.

\subsubsection*{Two dimensions}

We first focus on the bidimensional case ($d=2$). The first class has a fixed mean and covariance, while the second class evolves smoothly in both mean and covariance to produce a variety of configurations. In particular, the first class is defined as follows:

\[\mu_0= (0,0), \, \Sigma_0=
\begin{pmatrix}
	5.000000 & 2.181564 \\
	2.181564 & 5.000000
\end{pmatrix}.\]

To generate the covariance matrix for the first class, we construct a correlation matrix $R$ with diagonal entries equal to $1$ and off-diagonal entries initialized at $0.4$, to which symmetric perturbations from $\mathcal{N}(0,0.1)$ are added to enhance realism. The matrix is then adjusted to be positive definite by correcting negative eigenvalues and subsequently normalized to preserve ones on the diagonal. Setting the variances of both variables to $5$, we define a diagonal matrix $D$ with their square roots and obtain the covariance matrix as $\Sigma_0 = D R D$. This procedure generates a fixed, slightly inclined elliptical scatter plot.

For the second class, the mean evolves over time. Specifically, it is defined as
\[
\mu_{1,t} = \mu_0 + t \cdot (\mu_{1,1} - \mu_0),
\]
which corresponds to a linear interpolation from the initial point \(\mu_{0}=(0,0)\) to $\mu_{1,1}=(10,10)$ as \(t\) increases. As a result, the mean of the second class gradually shifts along the diagonal direction. 
On the other hand, the initial covariance is given by
\[
\Lambda = \begin{pmatrix}1 & 0 \\ 0 & 15\end{pmatrix},
\] 
and is rotated over time using the angle
\[
\theta_t = (1 - t) \cdot 0.8 \pi,
\] 
so that at \(t=0\) the rotation is maximal (approximately \(144^\circ\)) and at \(t=1\) there is no rotation. The rotated covariance matrix is then 
\[
\Sigma_{1,\text{rotated}} = R_t \Lambda R_t^\top,
\] 
where \(R_t\) is the 2D rotation matrix for angle \(\theta_t\), resulting in a rotated ellipse. Finally, the covariance of the second class is smoothly interpolated toward that of the first class according to
\[
\Sigma_{1,t} = (1 - t) \Sigma_{1,\text{rotated}} + t \Sigma_0,
\] 
ensuring a gradual transition from the rotated shape of the second class to the shape of the first class.\footnote{Note that \(\Sigma_{1,t}\) remains symmetric and positive definite, as both \(\Lambda\) and \(\Sigma_0\) are symmetric positive definite.}

\subsubsection*{Three dimensions}

When $d=3$, the data is generated following procedures analogous to those used in the two-dimensional setting: one class remains fixed while the other varies with~$t$. 
The first and fixed class is defined with mean vector $\mu_0 = (0,0,0)$ and covariance matrix
\[
\Sigma_0 =
\begin{pmatrix}
	5.000000 & 2.316431 & 2.464828 \\
	2.316431 & 5.000000 & 1.746522 \\
	2.464828 & 1.746522 & 4.000000
\end{pmatrix}.
\]
In this case, $D=\mathrm{diag}(\sqrt{5},\sqrt{5},\sqrt{4})$ and the remainder of the process follows the same logic as in the two-dimensional case.  
For the second class, the shifting mean is:
\[
\mu_{1,t} = \mu_0 + t \cdot (\mu_{1,1} - \mu_0),
\]
where $\mu_{1,1} = (10,10,10)$. Hence, the mean interpolates linearly from $(0,0,0)$ to $(10,10,10)$ as $t$ increases.
On the other hand, the initial covariance matrix is given by
\[
\Lambda =
\begin{pmatrix}
	1 & 0 & 0 \\
	0 & 15 & 0 \\
	0 & 0 & 11
\end{pmatrix},
\]
generated randomly and subsequently rotated around the $Z$-axis.

\subsubsection*{Ten dimensions}

When $d=10$, again the data is generated following procedures analogous to those used in the two-dimensional setting: one class remains fixed while the other varies with~$t$.

The first and fixed class is defined with mean vector $\mu_0 = (0,0,\dots,0) \in \mathbb{R}^{10}$. The covariance matrix is
{\footnotesize
	\[
	\Sigma_0 =
	\begin{pmatrix}
		5.00 & 1.85 & -0.39 & 0.50 & 0.19 & 0.58 & -0.70 & -2.96 & 0.96 & 0.40 \\
		1.85 & 5.00 & -2.26 & 0.78 & -0.32 & -1.41 & 0.35 & -0.26 & 0.16 & -0.14 \\
		-0.39 & -2.26 & 4.00 & 1.02 & 0.61 & 2.53 & 0.99 & 1.58 & 0.05 & 0.17 \\
		0.50 & 0.78 & 1.02 & 3.00 & -0.51 & 0.89 & 2.07 & -2.10 & -0.06 & 0.31 \\
		0.19 & -0.32 & 0.61 & -0.51 & 2.00 & 0.10 & -0.88 & -0.98 & -0.48 & -0.20 \\
		0.58 & -1.41 & 2.53 & 0.89 & 0.10 & 8.00 & 3.14 & 2.09 & 0.49 & -0.31 \\
		-0.70 & 0.35 & 0.99 & 2.07 & -0.88 & 3.14 & 9.00 & 2.93 & -0.19 & -0.43 \\
		-2.96 & -0.26 & 1.58 & -2.10 & -0.98 & 2.09 & 2.93 & 20.00 & -0.23 & -0.83 \\
		0.96 & 0.16 & 0.05 & -0.06 & -0.48 & 0.49 & -0.19 & -0.23 & 1.00 & 0.01 \\
		0.40 & -0.14 & 0.17 & 0.31 & -0.20 & -0.31 & -0.43 & -0.83 & 0.01 & 0.20 \\
	\end{pmatrix}.
	\]
}
Unlike the two- and three-dimensional cases, the correlations among variables are set to zero and perturbations are introduced following a normal distribution with mean 0 and standard deviation 0.5, reflecting the expectation of both positive and negative correlations in higher dimensions.  

For the second class, the shifting mean is:
\[
\mu_{1,t} = \mu_0 + t \cdot (\mu_{1,1} - \mu_0),
\]
where $\mu_{1,1}=(10,10,\dots,10) \in \mathbb{R}^{10}$.
The initial covariance matrix is
\[
\Lambda=
\begin{pmatrix}
	1 & 0 & 0 & 0 & 0 & 0 & 0 & 0 & 0 & 0 \\
	0 & 15 & 0 & 0 & 0 & 0 & 0 & 0 & 0 & 0 \\
	0 & 0 & 11 & 0 & 0 & 0 & 0 & 0 & 0 & 0 \\
	0 & 0 & 0 & 13 & 0 & 0 & 0 & 0 & 0 & 0 \\
	0 & 0 & 0 & 0 & 15 & 0 & 0 & 0 & 0 & 0 \\
	0 & 0 & 0 & 0 & 0 & 11 & 0 & 0 & 0 & 0 \\
	0 & 0 & 0 & 0 & 0 & 0 & 8 & 0 & 0 & 0 \\
	0 & 0 & 0 & 0 & 0 & 0 & 0 & 4 & 0 & 0 \\
	0 & 0 & 0 & 0 & 0 & 0 & 0 & 0 & 14 & 0 \\
	0 & 0 & 0 & 0 & 0 & 0 & 0 & 0 & 0 & 14 \\
\end{pmatrix},
\]
generated randomly and rotations are applied in the $(4,7)$-plane relative to dimension~8.

\section{Additional figures for geometrically structured data}\label{Section:figures-geometric}

This section provides extended visualizations of classification results on geometrically structured data, complementing the figures presented in the main text. We include example datasets for each configuration of intersecting planes, spiral–plane structures, and tori. Misclassified points are highlighted to illustrate the regions where each classifier (MST-Class, MST-RClass, and KNN) faces the greatest challenges. These figures allow for a more detailed assessment of classifier performance across the different geometric scenarios.

\subsubsection*{Intersecting planes}

\begin{figure}[!htbp]
	\centering
	
	% --------- Subfigure 1 ---------
	\begin{subfigure}[b]{0.9\textwidth}
		\centering
		\begin{tikzpicture}
			% Imagen
			\node[anchor=south west,inner sep=0] (img) at (0,0)
			{\includegraphics[width=\textwidth]{planes_config1.png}};
			
			% Títulos MST-Class, KNN, MST-RClass + título de configuración
			\begin{scope}[x={(img.south east)},y={(img.north west)}, every node/.style={font=\footnotesize, text centered}]
				% Títulos sobre cada panel
				\node at (0.19,0.8) {\footnotesize  MST-Class};
				\node at (0.525,0.8) {\footnotesize MST-RClass};
				\node at (0.86,0.8) {\footnotesize KNN};
				% Título principal de la subfigura
				\node at (0.51,0.9) {\textbf{Configuration 1}};
			\end{scope}
		\end{tikzpicture}
		\label{fig:planes_config1}
	\end{subfigure}
	\vspace{-2cm}
	
	% --------- Subfigure 2 ---------
	\begin{subfigure}[b]{0.9\textwidth}
		\centering
		\begin{tikzpicture}
			\node[anchor=south west,inner sep=0] (img) at (0,0)
			{\includegraphics[width=\textwidth]{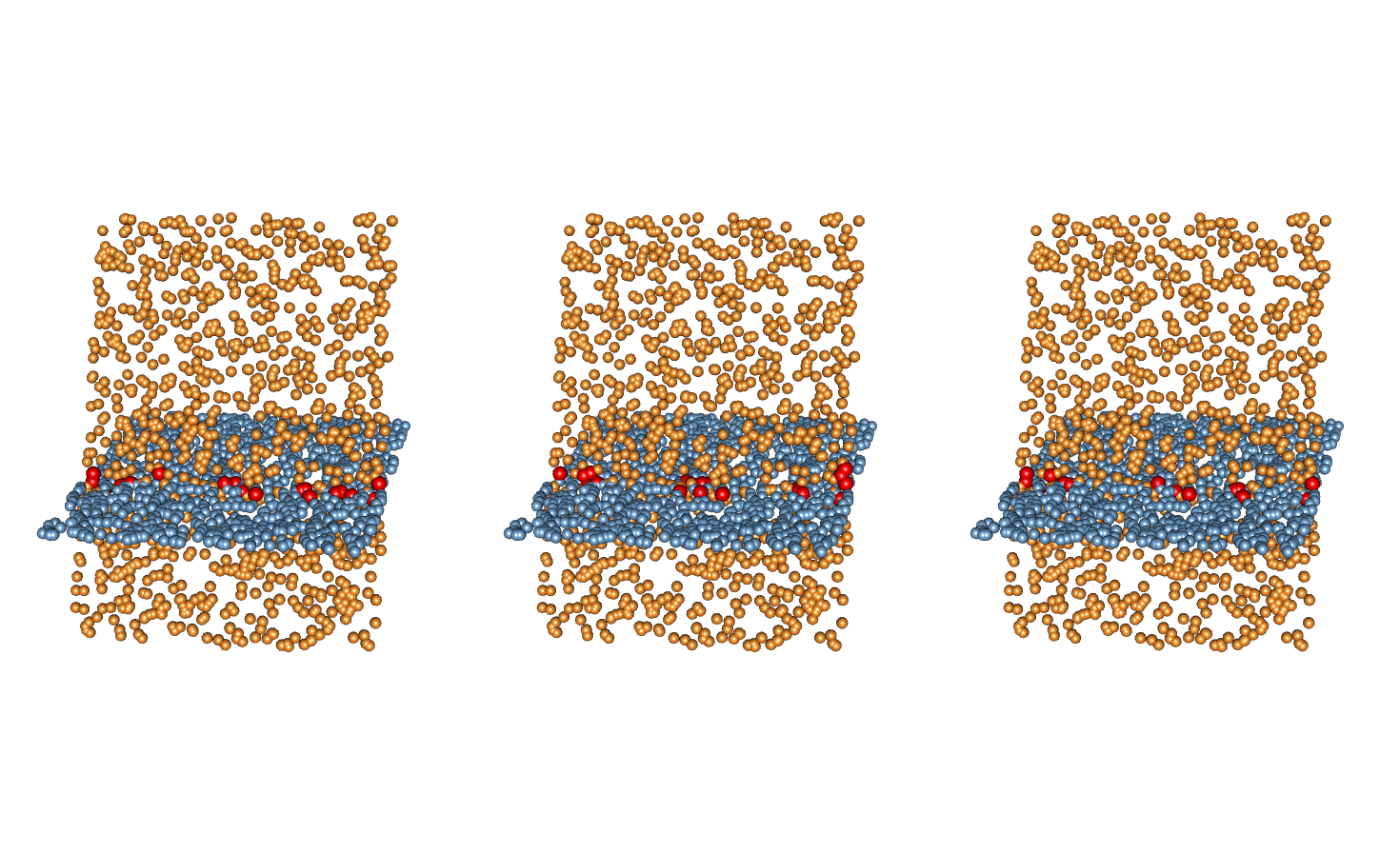}};
			\begin{scope}[x={(img.south east)},y={(img.north west)}, every node/.style={font=\footnotesize, text centered}]
				\node at (0.19,0.8) {MST-Class};
				\node at (0.525,0.8) {MST-RClass};
				\node at (0.86,0.8) {KNN};
				\node at (0.51,0.9) {\textbf{Configuration 2}};
			\end{scope}
		\end{tikzpicture}
		\label{fig:planes_config2}
	\end{subfigure}
	
	\vspace{-2cm}
	% --------- Subfigure 3 ---------
	\begin{subfigure}[b]{0.9\textwidth}
		\centering
		\begin{tikzpicture}
			\node[anchor=south west,inner sep=0] (img) at (0,0)
			{\includegraphics[width=\textwidth]{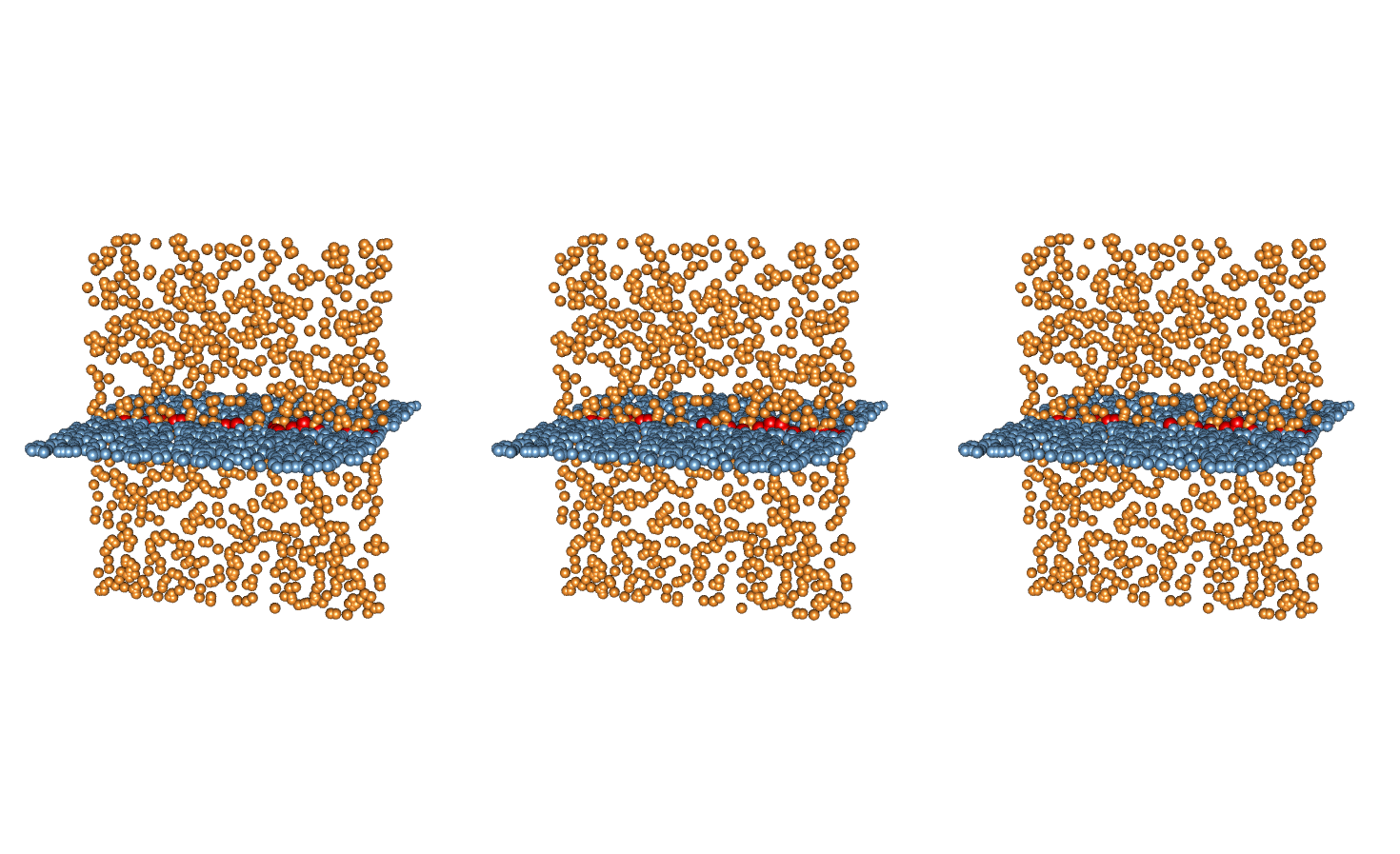}};
			\begin{scope}[x={(img.south east)},y={(img.north west)}, every node/.style={font=\footnotesize, text centered}]
				\node at (0.19,0.78) {MST-Class};
				\node at (0.525,0.78) {MST-RClass};
				\node at (0.86,0.78) {KNN};
				\node at (0.51,0.88) {\textbf{Configuration 3}};
			\end{scope}
		\end{tikzpicture}
		\label{fig:planes_config3}
	\end{subfigure}
	
	\vspace{-1.5cm}
	\caption{In red, misclassified points for the three configurations of intersecting planes.}
	\label{fig:planes_all_configs}
\end{figure}

Figure~\ref{fig:planes_all_configs} shows classification results for example datasets on two intersecting planes in $\mathbb{R}^3$. Each dataset is generated by sampling points on the planes, with the second plane rotated and vertically shifted according to the three configurations discussed in the main text. As expected, the points that are challenging for KNN largely coincide with those misclassified by MST-Class and MST-RClass, and are primarily located in the intersecting regions of the planes. Similar patterns are observed across all three configurations. 

Recall that we had concluded that MST-Class and KNN exhibit comparable accuracy across all configurations, with slightly higher values than those achieved by MST-RClass.

\subsubsection*{Intersecting spiral-plane}

\begin{figure}[!htbp]
	\centering
	
	% --------- Subfigure 1 ---------
	\begin{subfigure}[b]{0.9\textwidth}
		\centering
		\begin{tikzpicture}
			% Imagen
			\node[anchor=south west,inner sep=0] (img) at (0,0)
			{\includegraphics[width=\textwidth]{swissroll_config1.png}};
			
			% Títulos MST-Class, KNN, MST-RClass + título de configuración
			\begin{scope}[x={(img.south east)},y={(img.north west)}, every node/.style={font=\footnotesize, text centered}]
				% Títulos sobre cada panel
			\node at (0.18,0.8) {\footnotesize  MST-Class};
			\node at (0.51,0.8) {\footnotesize MST-RClass};
			\node at (0.84,0.8) {\footnotesize KNN};
				% Título principal de la subfigura
				\node at (0.51,0.9) {\textbf{Configuration 1}};
			\end{scope}
		\end{tikzpicture}
		\label{fig:spiral_plate_config1}
	\end{subfigure}
	\vspace{-2cm}
	
	% --------- Subfigure 2 ---------
	\begin{subfigure}[b]{0.9\textwidth}
		\centering
		\begin{tikzpicture}
			\node[anchor=south west,inner sep=0] (img) at (0,0)
			{\includegraphics[width=\textwidth]{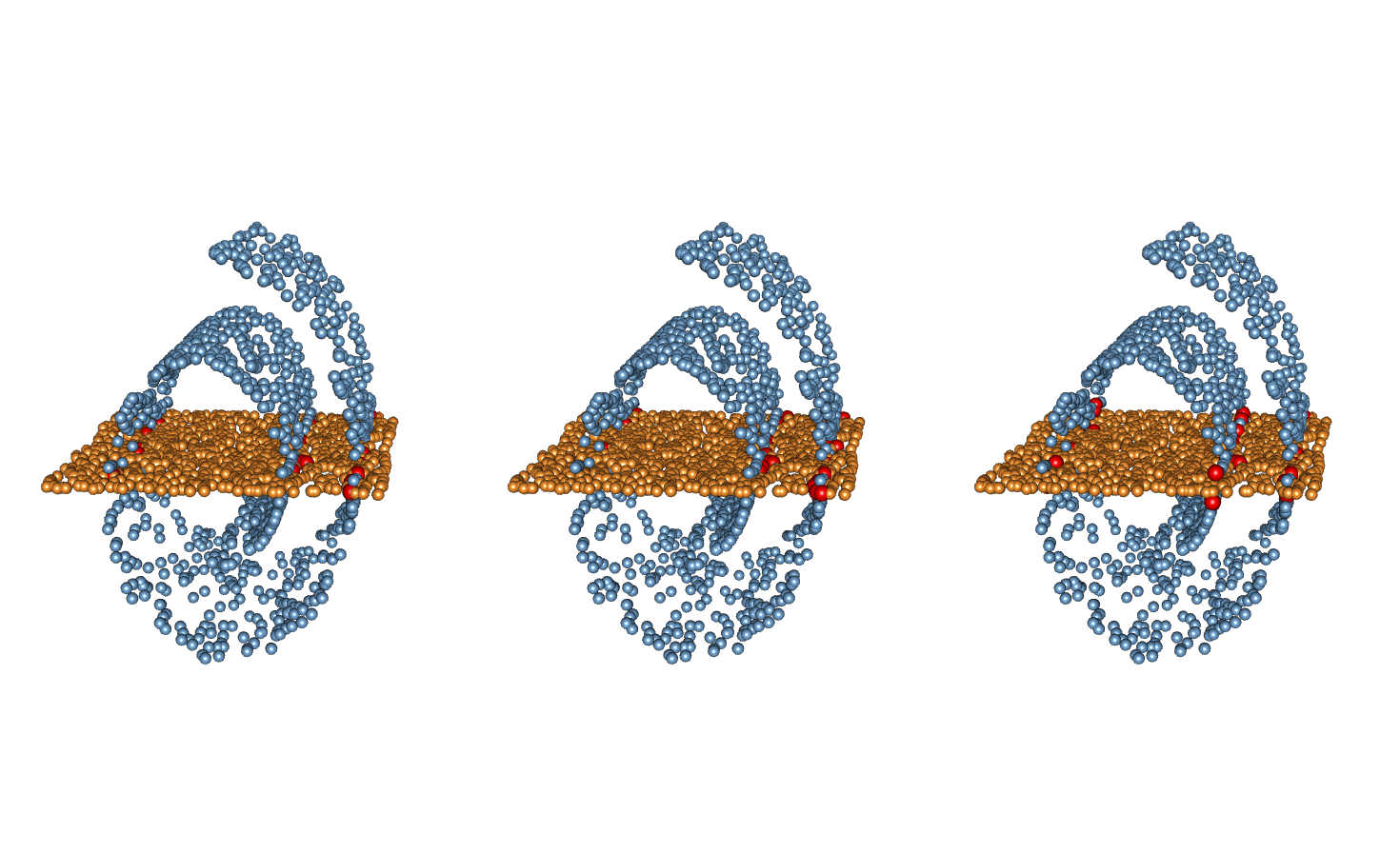}};
			\begin{scope}[x={(img.south east)},y={(img.north west)}, every node/.style={font=\footnotesize, text centered}]
			\node at (0.18,0.8) {\footnotesize  MST-Class};
\node at (0.51,0.8) {\footnotesize MST-RClass}; 
\node at (0.84,0.8) {\footnotesize KNN};
				\node at (0.51,0.9) {\textbf{Configuration 2}};
			\end{scope}
		\end{tikzpicture}
		\label{fig:spiral_plate_config2}
	\end{subfigure}
	
		\vspace{-2cm}
	% --------- Subfigure 3 ---------
	\begin{subfigure}[b]{0.9\textwidth}
		\centering
		\begin{tikzpicture}
			\node[anchor=south west,inner sep=0] (img) at (0,0)
			{\includegraphics[width=\textwidth]{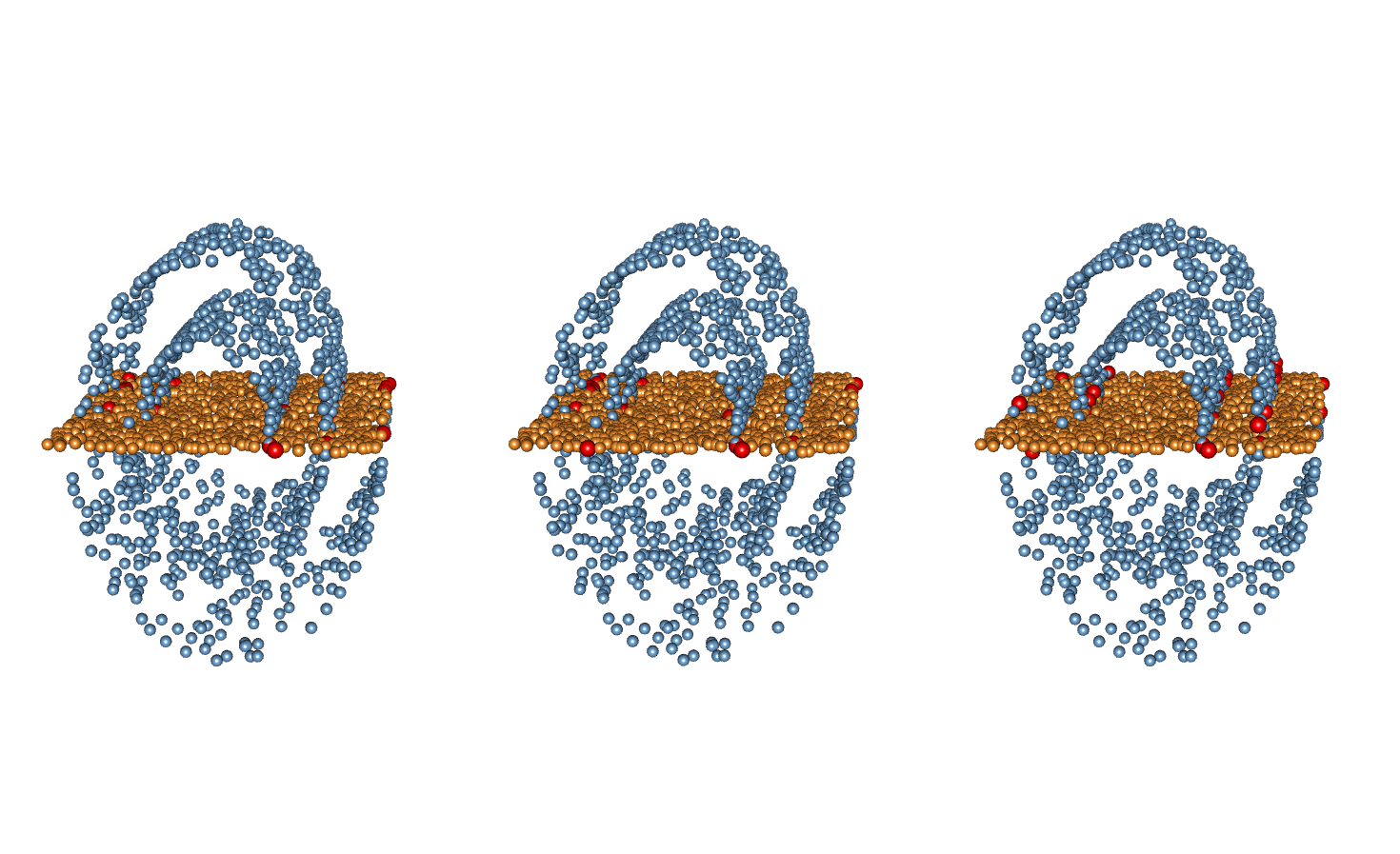}};
			\begin{scope}[x={(img.south east)},y={(img.north west)}, every node/.style={font=\footnotesize, text centered}]
			\node at (0.18,0.8) {\footnotesize  MST-Class};
\node at (0.51,0.8) {\footnotesize MST-RClass};
\node at (0.84,0.8) {\footnotesize KNN};
				\node at (0.51,0.9) {\textbf{Configuration 3}};
			\end{scope}
		\end{tikzpicture}
		\label{fig:spiral_plate_config3}
	\end{subfigure}
	
	\vspace{-1.5cm}
	\caption{In red, misclassified points for the three configurations of spiral-plane.}
\label{fig:spiral_plate_all_configs}
\end{figure}

Figure~\ref{fig:spiral_plate_all_configs} shows classification results for example datasets on intersecting spiral–plane structures in $\mathbb{R}^3$. Each dataset corresponds to one of three configurations, varying the angular range and spiral radius, as described in the main paper. As mentioned there, similar to the intersecting planes, the points that are challenging for KNN largely coincide with those misclassified by MST-Class and MST-RClass, with most errors concentrated near the class boundaries. 

Recall that MST-Class generally attains slightly higher accuracy than MST-RClass across all configurations. However, these differences are minor, and MST-RClass remains competitive, with performance comparable to that of KNN. The misclassified points highlighted in these figures illustrate and support these overall performance patterns.

\subsubsection*{Intersecting tori}

\begin{figure}[!htbp]
	\centering
	
	% --------- Subfigure 1 ---------
	\begin{subfigure}[b]{0.9\textwidth}
		\centering
		\begin{tikzpicture}
			% Imagen
			\node[anchor=south west,inner sep=0] (img) at (0,0)
			{\includegraphics[width=\textwidth]{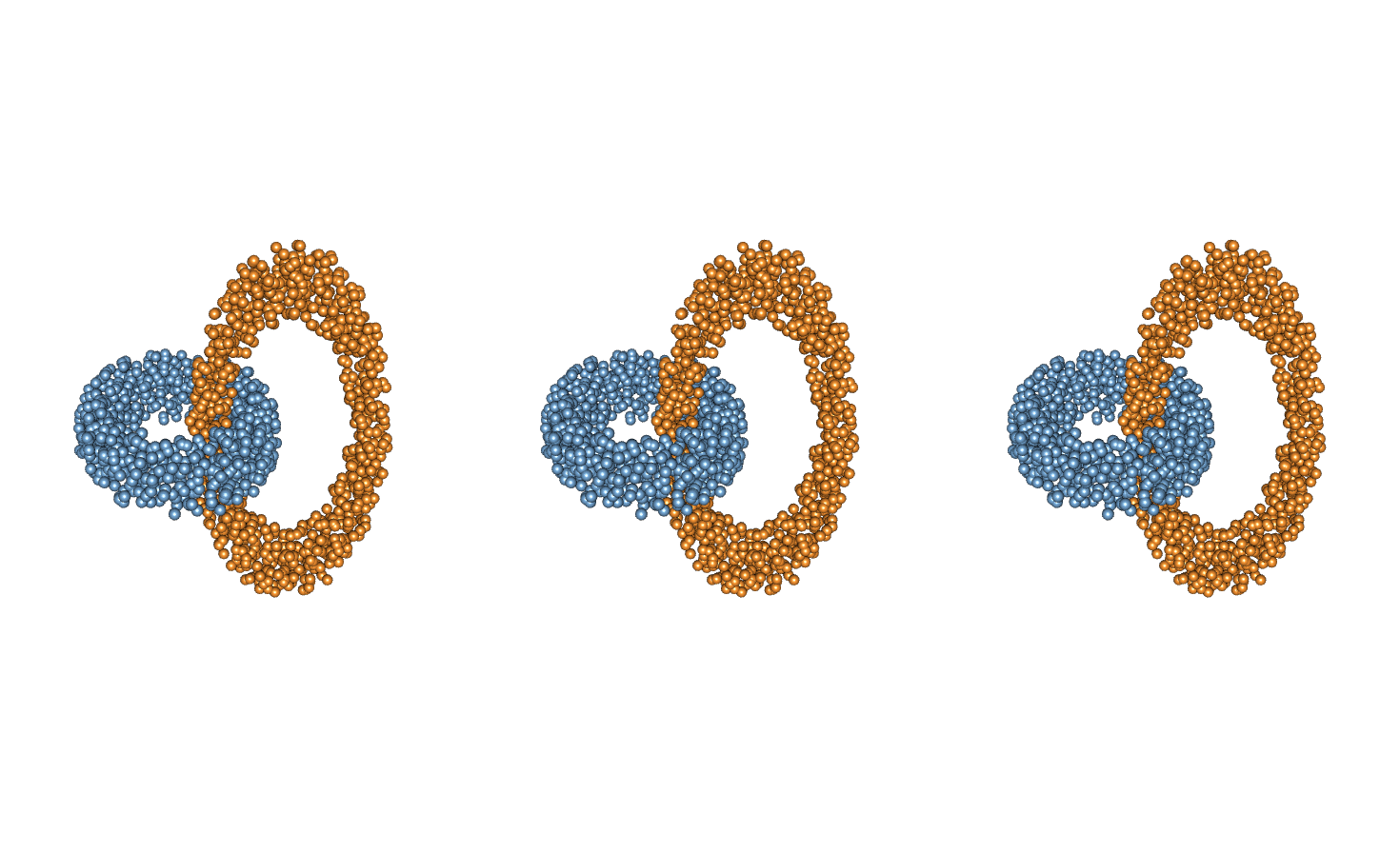}};
			
			% Títulos MST-Class, KNN, MST-RClass + título de configuración
			\begin{scope}[x={(img.south east)},y={(img.north west)}, every node/.style={font=\footnotesize, text centered}]
				% Títulos sobre cada panel
				\node at (0.17,0.8) {\footnotesize  MST-Class};
				\node at (0.50,0.8) {\footnotesize MST-RClass};
				\node at (0.83,0.8) {\footnotesize KNN};
				% Título principal de la subfigura
				\node at (0.51,0.9) {\textbf{Configuration 1}};
			\end{scope}
		\end{tikzpicture}
		\label{fig:torus_config1}
	\end{subfigure}
	
	\vspace{-2cm}
	
	% --------- Subfigure 2 ---------
	\begin{subfigure}[b]{0.9\textwidth}
		\centering
		\begin{tikzpicture}
			\node[anchor=south west,inner sep=0] (img) at (0,0)
			{\includegraphics[width=\textwidth]{torus_config2.png}};
			\begin{scope}[x={(img.south east)},y={(img.north west)}, every node/.style={font=\footnotesize, text centered}]
				\node at (0.17,0.8) {\footnotesize  MST-Class};
				\node at (0.50,0.8){\footnotesize MST-RClass}; 
				\node at (0.83,0.8) {\footnotesize KNN};
				\node at (0.51,0.9) {\textbf{Configuration 2}};
			\end{scope}
		\end{tikzpicture}
		\label{fig:torus_config2}
	\end{subfigure}
		\vspace{-2cm}
	
	% --------- Subfigure 3 ---------
	\begin{subfigure}[b]{0.9\textwidth}
		\centering
		\begin{tikzpicture}
			\node[anchor=south west,inner sep=0] (img) at (0,0)
			{\includegraphics[width=\textwidth]{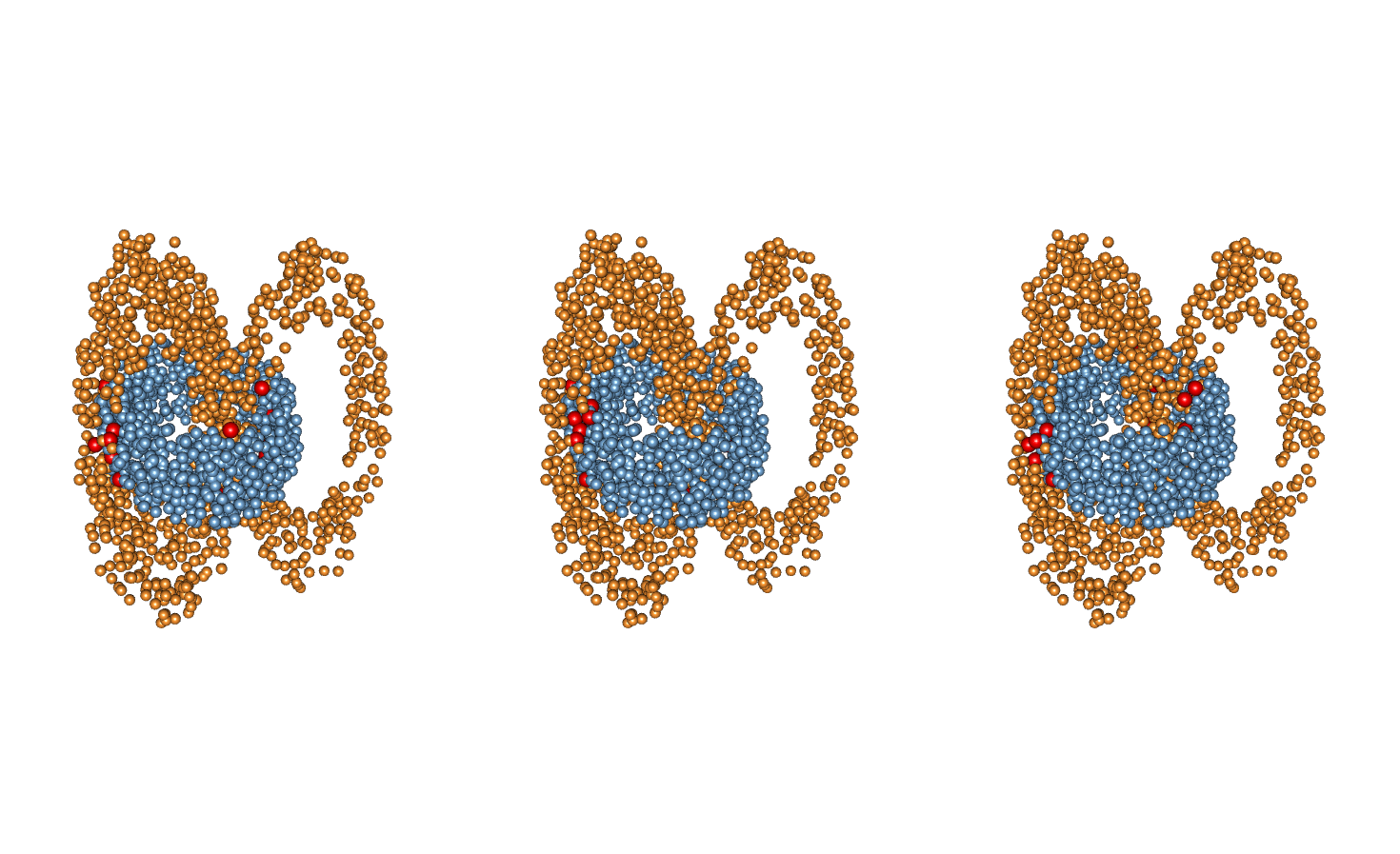}};
			\begin{scope}[x={(img.south east)},y={(img.north west)}, every node/.style={font=\footnotesize, text centered}]
				\node at (0.17,0.8) {\footnotesize  MST-Class};
				\node at (0.50,0.8) {\footnotesize MST-RClass};
				\node at (0.83,0.8) {\footnotesize KNN};
				\node at (0.51,0.9) {\textbf{Configuration 3}};
			\end{scope}
		\end{tikzpicture}
	\label{fig:torus_config3}
	\end{subfigure}
	
	\vspace{-1.5cm}
\caption{In red, misclassified points for the three configurations of tori.}
\label{fig:torus_all_configs}
\end{figure}

Figure \ref{fig:torus_all_configs} shows classification results for example datasets on torus-shaped class structures in $\mathbb{R}^3$. Each dataset corresponds to one of three configurations, varying the number, size, and placement of the tori, as described in the main text. Note that in Configuration 1 there are no red points, since, as reported, the mean accuracy for all methods under this configuration was 1. 

As previously discussed, all three classifiers (MST-Class, MST-RClass, and KNN) achieve very high accuracy, often reaching perfect classification. In Configurations 2 and 3, MST-Class and MST-RClass show a slight improvement over KNN.

\section{A case study: aircraft trajectories}

In this section, we present the classification performance of MST-Class, MST-RClass, and KNN on real aircraft trajectory data from the TrajAir dataset \citep{trajectories}. Thirty pairs of trajectories were randomly selected, with each pair treated as a binary classification problem and points labeled according to the aircraft to which they belong. 

Table~\ref{tab:aircraft-random} reports the accuracies obtained by each method, with MST-Class and MST-RClass values that are equal to or higher than the corresponding KNN accuracy highlighted in bold. In this relatively straightforward scenario, MST-Class matches or outperforms KNN in 24 out of 30 cases, while MST-RClass does so in 16 cases. Overall, all methods achieve very high accuracy, and the high mean values of the parameter $h$ suggest that cross-validation successfully identifies the need for relatively large subsamples to capture the global trajectory structure.

\begin{table}[!htbp]
	\centering
	\resizebox{\textwidth}{!}{%
		\begin{tabular}{c c cc cc}
			\toprule
			Trajectory pair & MST-Class & \multicolumn{2}{c}{MST-RClass} & \multicolumn{2}{c}{KNN} \\
			\cmidrule(lr){3-4} \cmidrule(lr){5-6}
			&  & Accuracy & $h$ & Accuracy & $k$ \\
			\midrule
			1 & \textbf{1.000 (0.000)} & 0.999 (0.002) & 8.000 (2.582) & 1.000 (0.000) & 1.200 (0.633) \\
			2 & \textbf{1.000 (0.000)} & \textbf{0.999 (0.001)} & 9.500 (1.581) & 0.999 (0.001) & 10.400 (3.748) \\
			3 & \textbf{1.000 (0.000)} & 0.997 (0.002) & 10.000 (0.000) & 1.000 (0.000) & 1.000 (0.000) \\
			4 & 0.989 (0.008) & 0.990 (0.010) & 10.000 (0.000) & 0.991 (0.002) & 1.300 (0.483) \\
			5 & 0.985 (0.008) & 0.988 (0.007) & 6.000 (2.108) & 0.992 (0.003) & 1.800 (1.229) \\
			6 & \textbf{0.995 (0.006)} & \textbf{0.991 (0.008)} & 10.000 (0.000) & 0.982 (0.008) & 2.700 (0.483) \\
			7 & \textbf{1.000 (0.000)} & \textbf{0.999 (0.001)} & 8.000 (2.582) & 0.998 (0.004) & 2.100 (1.853) \\
			8 & 0.993 (0.011) & 0.981 (0.010) & 6.400 (4.648) & 1.000 (0.000) & 10.300 (2.983) \\
			9 & \textbf{1.000 (0.000)} & 0.999 (0.003) & 9.500 (1.581) & 0.999 (0.002) & 2.800 (2.700) \\
			10 & \textbf{1.000 (0.000)} & \textbf{1.000 (0.000)} & 10.000 (0.000) & 0.994 (0.005) & 1.200 (0.422) \\
			11 & \textbf{0.997 (0.005)} & \textbf{0.994 (0.006)} & 9.000 (2.108) & 0.990 (0.005) & 1.400 (0.843) \\
			12 & \textbf{1.000 (0.000)} & 0.996 (0.007) & 7.700 (3.093) & 1.000 (0.000) & 9.200 (3.853) \\
			13 & \textbf{1.000 (0.000)} & \textbf{1.000 (0.000)} & 5.500 (1.581) & 1.000 (0.000) & 27.000 (0.000) \\
			14 & \textbf{0.993 (0.006)} & \textbf{0.980 (0.008)} & 10.000 (0.000) & 0.962 (0.014) & 1.100 (0.316) \\
			15 & \textbf{0.998 (0.003)} & \textbf{0.994 (0.003)} & 8.500 (2.415) & 0.987 (0.007) & 2.700 (1.059) \\
			16 & \textbf{1.000 (0.000)} & \textbf{1.000 (0.000)} & 5.500 (1.581) & 0.999 (0.001) & 3.500 (1.780) \\
			17 & \textbf{0.983 (0.010)} & \textbf{0.984 (0.005)} & 10.000 (0.000) & 0.969 (0.006) & 3.600 (1.350) \\
			18 & \textbf{1.000 (0.000)} & 0.998 (0.004) & 7.500 (2.635) & 1.000 (0.000) & 6.300 (2.214) \\
			19 & \textbf{0.995 (0.005)} & \textbf{0.989 (0.009)} & 10.000 (0.000) & 0.974 (0.010) & 1.000 (0.000) \\
			20 & 0.989 (0.007) & \textbf{0.983 (0.009)} & 9.000 (2.108) & 0.974 (0.008) & 1.600 (0.699) \\
			21 & \textbf{0.999 (0.002)} & \textbf{0.997 (0.003)} & 8.500 (2.415) & 0.994 (0.003) & 1.400 (0.699) \\
			22 & \textbf{1.000 (0.000)} & 0.998 (0.005) & 8.000 (2.582) & 0.999 (0.002) & 1.000 (0.000) \\
			23 & \textbf{1.000 (0.000)} & \textbf{1.000 (0.000)} & 9.000 (2.108) & 0.994 (0.003) & 2.000 (0.943) \\
			24 & \textbf{1.000 (0.000)} & 0.997 (0.003) & 9.500 (1.581) & 0.998 (0.003) & 2.200 (1.874) \\
			25 & 0.997 (0.004) & 0.989 (0.008) & 9.500 (1.581) & 0.999 (0.001) & 2.200 (1.619) \\
			26 & \textbf{1.000 (0.000)} & 0.999 (0.001) & 8.500 (2.415) & 1.000 (0.000) & 13.000 (2.404) \\
			27 & 0.999 (0.002) & 0.999 (0.002) & 5.800 (3.967) & 1.000 (0.000) & 27.000 (0.000) \\
			28 & \textbf{1.000 (0.000)} & \textbf{1.000 (0.000)} & 10.000 (0.000) & 0.998 (0.002) & 1.600 (0.966) \\
			29 & \textbf{1.000 (0.000)} & \textbf{1.000 (0.000)} & 5.000 (0.000) & 1.000 (0.000) & 22.800 (3.853) \\
			30 & 0.996 (0.007) & 0.991 (0.009) & 10.000 (0.000) & 0.998 (0.004) & 1.000 (0.000) \\
			\bottomrule
		\end{tabular}
	}
	\caption{Classification results for \textit{MST-Class}, \textit{MST-RClass} and \textit{KNN} across 30 randomly selected trajectory pairs.}
	\label{tab:aircraft-random}
\end{table}

\bibliographystyle{elsarticle-harv} 
\bibliography{biblio}